\documentclass[letterpaper]{article} % DO NOT CHANGE THIS
\usepackage{style_sheets/acl}  % DO NOT CHANGE THIS

\usepackage{graphicx} % Required for inserting images
\usepackage{subcaption}
\usepackage{amsmath} 

\usepackage{multirow} % For cells spanning multiple rows in tables
\usepackage{makecell} % For customizing cell formats
\usepackage{natbib} % For citation management
\usepackage{pifont} % For special symbols like check marks and X marks
\usepackage{xcolor} % For defining custom colors
\usepackage[utf8]{inputenc} % For input encoding

\usepackage{tcolorbox}
\usepackage{booktabs}       % professional-quality tables
\usepackage{amsfonts}       % blackboard math symbols
\usepackage{nicefrac}       % compact symbols for 1/2, etc.
\usepackage{microtype}      % microtypography
\usepackage{xcolor}         % colors
\usepackage{tikz}
\usepackage{graphicx} % Required for adjusting font size and table resizing
\usepackage{array} % Required for custom column alignment
\usepackage{booktabs} % For better table lines

\newcommand{\benchmark}{MAGMA Benchmark}

\title{Are Large-Language Models Graph Algorithmic Reasoners?}

\author{
  Alexander K. Taylor, Anthony Cuturrufo, Vishal Yathish, Mingyu Derek Ma, \\
  \textbf{Wei Wang} \\
  University of California, Los Angeles \\
  \texttt{\{aktaylor, acc, vishalyathish1, ma, weiwang\}@cs.ucla.edu}
}

\begin{document}

\maketitle
\begin{abstract}
We seek to address a core challenge facing current Large Language Models (LLMs). LLMs have demonstrated superior performance in many tasks, yet continue to struggle with reasoning problems on explicit graphs that require multiple steps. We introduce a novel benchmark designed to evaluate LLM performance on classical algorithmic reasoning tasks on explicit graphs to address this gap. Our benchmark encompasses five fundamental algorithms: Breadth-First Search (BFS) and Depth-First Search (DFS) for connectivity, Dijkstra's algorithm and Floyd-Warshall algorithm for all nodes shortest path, and Prim's Minimum Spanning Tree (MST-Prim's) algorithm. Through extensive experimentation, we assess the capabilities of state-of-the-art LLMs in executing these algorithms step-by-step and systematically evaluate their performance at each stage. Our findings highlight the persistent challenges LLMs face in this domain and underscore the necessity for advanced prompting techniques and algorithmic instruction to enhance their graph reasoning abilities. This work presents MAGMA, the first comprehensive benchmark focused on LLMs completing classical graph algorithms, and provides a critical step toward understanding and improving their structured problem-solving skills (Code: \url{https://github.com/ataylor24/MAGMA}).
\end{abstract}

\section{Introduction}

In recent years, large language models (LLMs) have demonstrated remarkable capabilities across a wide range of natural language processing tasks. 
Despite these advances, LLMs continue to exhibit significant limitations in solving multistep reasoning problems as the available context increases. These limitations necessitate the development of problem sets designed to help LLMs learn to solve problems correctly across multiple solution steps.

\begin{figure}[h]
  \centering
\includegraphics[width=\columnwidth]{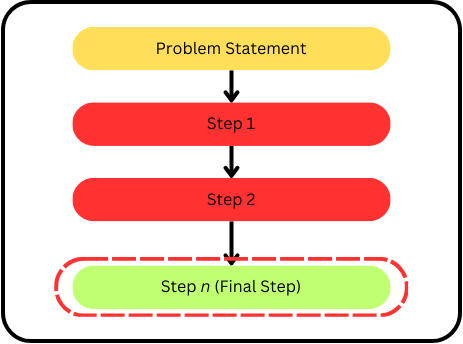}
  \caption{Illustration of a problem in which only the final step is solved correctly.}
\label{final_step_solution_insufficient}
\end{figure}

Graph-reasoning tasks provide a compelling area for evaluating multistep reasoning abilities due to their reliance on structured, ordered steps for accurate solutions.
Recent works exploring LLM performance on graph-reasoning tasks flatten multistep problems to only the problem statement and the final solution query, and use only the final step solution to evaluate the model, as exemplified in Figure \ref{final_step_solution_insufficient} \citep{liu2023evaluating, wang2023can, zhang2023llm4dyg, agrawal_exploring_2024, ge_graph_2024, mcleish2024, markeeva2024}. 
This simplification, while convenient for evaluation, overlooks the intermediate reasoning steps that reveal the model's understanding of each problem segment and its progression toward the final answer. For instance, when computing the shortest path in Dijkstra’s algorithm, intermediate steps involve checking and updating node distances based on edge weights. Focusing on these steps provides critical insights into the model’s comprehension and logical progression, enhancing both training and evaluation.

Training models to correctly identify and solve sub-problems has applications in fields such as AI for scientific discovery, where stepwise reasoning is essential. Therefore, exploring intermediate steps in graph reasoning tasks is valuable not only for assessing models’ problem-solving capacity, but also for applications requiring systematic and logical problem breakdowns.

In this work, we aim to demonstrate that incorporating intermediate solution steps into training and in-context reasoning benefits the performance of both fine-tuned and zero-shot LLMs on multistep graph reasoning problems.
To our knowledge, this is the first benchmark designed to fine-tune and evaluate LLMs' performance on each intermediate step within classical graph reasoning problems, highlighting their importance in effective model training and evaluation.

Classical graph algorithms, which require ordered, deterministic steps dictated by graph structure, are ideal for testing LLMs' ability to handle multistep reasoning. These algorithms have well-defined sequences that provide a structured framework for training and evaluating models on solution trajectories \citep{kleinbergtardos2005algorithm}. Additionally, algorithmic reasoning trajectories do \textit{not} have multiple equivalent solution paths, which allows for easily automated evaluation of intermediate solution steps.
Given these characteristics, classical graph algorithms are an ideal test bed for evaluating and enhancing the multistep reasoning capabilities of LLMs.
Prior works have shown that using intermediate solution steps, in addition to final solutions, can boost model performance \citep{deac2021neural, velickovic2021neural, ibarz2022generalist, numeroso2023dual, velickovic2022clrs}. 
However, these approaches have not yet fully explored graph reasoning problems in the context of LLMs, nor have they automated benchmark evaluations that use multistep trajectory data. Our work leverages the CLRS Benchmark to construct high-quality training and evaluation examples for selected algorithms, providing a novel and systematic way to test the multistep reasoning capabilities of LLMs \citep{velickovic2022clrs, ibarz2022generalist}.

We introduce a comprehensive benchmark designed to evaluate LLM performance on classical graph algorithms using intermediate steps.
We title our benchmark the \textit{Multistep AlgorithMic reAsoning Benchmark} (MAGMA).\footnote{We also pay homage to the CLRS Algorithmic Reasoning Benchmark \citep{velickovic2022clrs}}.
Our benchmark includes five fundamental algorithms- Breadth-first Search, Depth-first Search, Dijkstra, Floyd-Warshall, and Prim's Minimum Spanning Tree-that together provide a comprehensive basis for assessing LLMs’ graph reasoning skills.
Through systematic evaluations of current state-of-the-art LLMs, our benchmark offers valuable insights into the models’ current limitations, as well as prompting techniques that could further enhance their multistep reasoning abilities.

Our work makes several key contributions. 
First, we introduce the first benchmark focused on evaluating LLMs' ability to perform the intermediate steps of classical graph algorithms by adapting trajectories from the CLRS benchmark \citep{velickovic2022clrs}, a standard neural algorithmic reasoning benchmark.
Second, we conduct an extensive evaluation of the multistep algorithmic reasoning capabilities of LLMs, from which we observe the following: \begin{itemize}
    \item Incorporating intermediate steps into in-context learning and fine-tuning significantly improves algorithmic reasoning performance.
    \item Smaller fine-tuned models outperform larger foundation language models on graph algorithmic reasoning tasks.
    \item Models fine-tuned with intermediate steps demonstrate sensitivity to extraneous information, underscoring the need for clarity in step-by-step reasoning.
\end{itemize} 
Through these contributions, we aim to establish a foundation for future work to address and overcome the challenges observed in multistep reasoning within algorithmic tasks.

\section{Related Works}

\subsection{Algorithmic Reasoning}
Many prior works have explored the application of machine learning to classic computer science algorithms \cite{graves2016hybrid,xu2019can,reed2015neural,vinyals2015pointer,bello2016neural,kool2018attention,selsam2018learning,selsam2019guiding,yoon2018inference,bronstein2017geometric,hamilton2017representation,battaglia2018relational,li2015gated,kipf2016semi,gilmer2017neural,deac2021neural,velickovic2021neural,xhonneux2021how,ibarz2022generalist,numeroso2023dual,velickovic2022clrs}.
\cite{velickovic2021neural} introduced the use of intermediate steps in addition to ground-truth solutions to guide learning, both for ease of optimization and to constrain the freedom of the model in finding the solution. This led to multiple works using inter.mediate steps to improve algorithmic reasoning performance for both single and multi-algorithm task settings. \citep{deac2021neural,xhonneux2021how,numeroso2023dual}.
Lastly, the CLRS Algorithmic Reasoning benchmark was constructed for algorithmic reasoning on GNNs and established standards for GNN performance for 30 algorithms in single and multi-algorithm settings \cite{velickovic2022clrs, ibarz2022generalist}.
We adapt the graph and algorithmic trajectory generation from the CLRS benchmark for use in the \benchmark~to improve the reasoning capabilities of LLMs on the full set of solutions for graph problems and allow for future comparison of LLMs to GNNs on algorithmic reasoning.

\subsection{Graph Reasoning Tasks with LLMs}

Applying large language models to graph reasoning tasks is an emerging research area that has made significant strides in recent years \citet{li_survey_2024, zhao2023graphtext, guo2023gpt4graph, liu2023evaluating, wang2023can, zhang2023llm4dyg, agrawal_exploring_2024, ge_graph_2024}. 
We focus on works in which LLMs are used as \textit{predictors} in reasoning tasks involving set and graph operations, as defined in \citet{chen2023exploring}. 
Most approaches frame the objective as identifying paths or substructures within graphs and do not specify an algorithm or set of required steps to adhere to in the reasoning process \citep{liu2023evaluating, wang2023can, zhang2023llm4dyg, agrawal_exploring_2024, ge_graph_2024}. 
Prior works that \textit{do} include specific algorithms in their prompts only evaluate the final solution step \citep{wang_can_nodate, liu2023evaluating, ge_graph_2024, markeeva2024, mcleish2024}. 

These studies seek to address a known weakness in current state-of-the-art LLMs regarding algorithmic reasoning tasks on graph data.
However, these works formulate algorithmic reasoning as a one-step question-answering task, which does not consider the intermediate reasoning steps followed by an execution of the algorithm.
Our work aims to fill this gap by constructing step-by-step executions of each algorithm in a chat format, which provides training data better suited to multistep reasoning tasks.

\section{Benchmark Construction}
The \benchmark~ consists of step-by-step graph algorithm executions on explicit graphs in natural language for the following algorithms: BFS, DFS, Dijkstra, Floyd-Warshall, and MST-Prim.
We leverage the framework established in the CLRS benchmark to sample graphs and construct algorithm trajectories to translate into natural language.

\subsection{Adapting the CLRS Benchmark}
\subsubsection{Algorithm Selection}
We select a subset of the available graph algorithms from the CLRS Benchmark for inclusion in the \benchmark~. 
\begin{itemize}
    \item \textbf{Breadth-First Search}: standard example of a simple algorithm (for humans), and thus included in many algorithmic reasoning works \citep{velickovic2021neural,xhonneux2021how,ibarz2022generalist,velickovic2022clrs,liu2023evaluating, wang2023can, ge_graph_2024}.
    \item \textbf{Depth-First Search}: shown to be quite challenging for neural algorithmic reasoning approaches on the CLRS benchmark \citep{velickovic2021neural,xhonneux2021how,ibarz2022generalist,velickovic2022clrs,liu2023evaluating, wang2023can, ge_graph_2024}.
    \item \textbf{Dijkstra}: standard example of a shortest-path problem, and commonly included in both LLM graph reasoning benchmarks and neural algorithmic reasoning works \citep{ibarz2022generalist}.
    \item \textbf{Floyd-Warshall}: shortest-path problem that has consistently proven challenging for prior neural algorithmic reasoning approaches to perform \citep{ibarz2022generalist}.
    \item \textbf{Prim's MST}: included because, to the best of our knowledge, minimum spanning trees have not been thoroughly explored in prior LLM-graph algorithmic reasoning works \citep{ibarz2022generalist, mcleish2024, markeeva2024}.
\end{itemize}

\subsubsection{Graph Sampling}
We follow the CLRS Benchmark in using Erdos-Renyi graphs for our algorithm execution trajectories. We sample graphs of size $n \in\{[5,15] \cup \{20,50\}\}$ with edge probabilities of 0.5. For each algorithm, we enforce the uniqueness of each data point to prevent data leakage.

\subsubsection{Problem Trajectories}

For each problem trajectory, we construct the \textit{problem statement}, \textit{intermediate steps}, and \textit{final step output} using the \textit{inputs}, \textit{hints}, and \textit{outputs} from the CLRS Benchmark. The components of these trajectories are as follows:

\begin{itemize}
    \item \textbf{Problem Statement}: 
    We convert the adjacency and weight matrices into lexicographically ordered string edge lists, following the format of prior graph reasoning benchmarks \citep{liu2023evaluating, wang2023can, zhang2023llm4dyg, agrawal_exploring_2024, ge_graph_2024}. We also include the source node if it is provided in the CLRS input.
    
    \item \textbf{Intermediate Steps}: 
    The hints provided by the CLRS benchmark consist of the algorithm-specific components of the execution of each algorithm and include an intermediate solution \cite{velickovic2022clrs}.   
    We construct a chat representation of the algorithmic execution, alternating between an intermediate step prompt (user) and intermediate solution (assistant). We also include a setting in which we translate data from matrices (CLRS format) to natural language, as shown in Figure \ref{hints} 
    
    \item \textbf{Final step output}: We use the final step of the set of hints as the final step solution to maintain consistency with the intermediate solutions for that algorithm.
    
\end{itemize}

\begin{figure}[h]
    \centering  
    \input{./Figures/Examples/clrs_example} % Adjust the path as needed
    \caption{Illustration of splitting an algorithmic execution step into chat format.}
    \label{hints}
\end{figure}

\subsection{Prompt Reasoning Strategies}
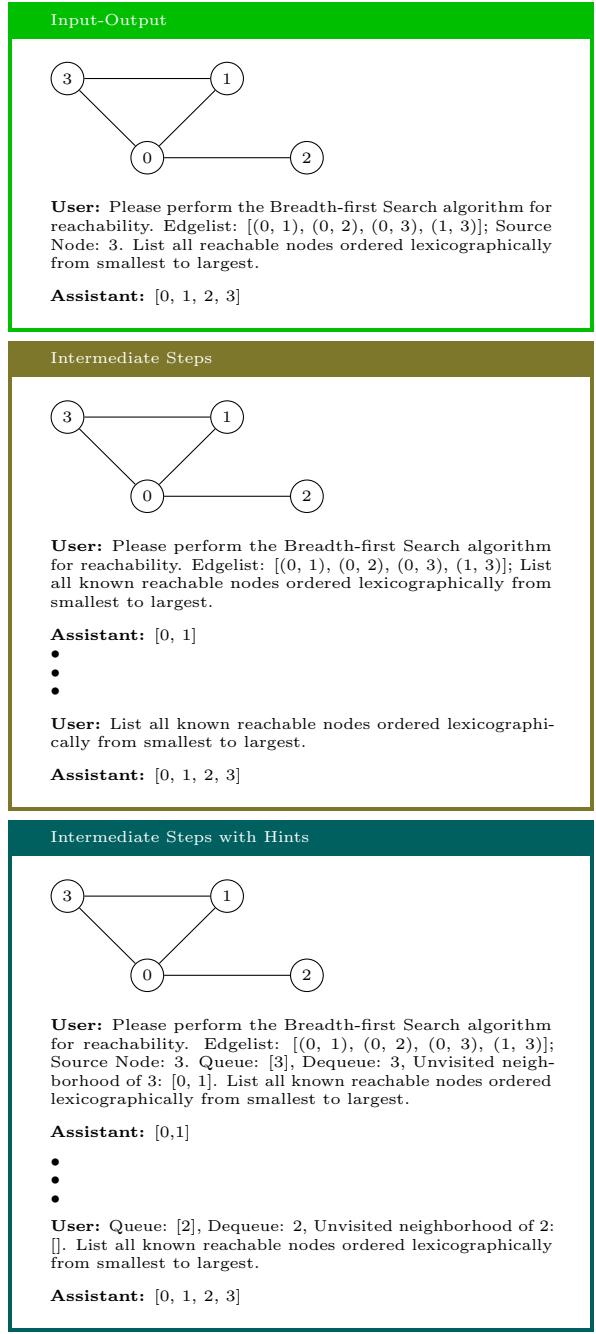
\begin{figure}[h!]
    \centering  
    {\tiny % Start group with even smaller font size
    \begin{tcolorbox}[colback=white!10!white,colframe=green!75!black,title= Input-Output,sharp corners, box align=top, width=\linewidth] % Set width to linewidth
        \begin{tikzpicture}[scale=0.7] % Adjust the scale here
            \node[circle, draw] (0) at (0,0) { 0};
            \node[circle, draw] (1) at (1.5,1.5) { 1};
            \node[circle, draw] (2) at (3,0) { 2};
            \node[circle, draw] (3) at (-1.5,1.5) { 3};
            \draw (0) -- node[above] {} (1);
            \draw (0) -- node[above] {} (2);
            \draw (0) -- node[left] {} (3);
            \draw (1) -- node[below] {} (3);
        \end{tikzpicture}
        \\[1em] % Ensure spacing between graphic and text
        \textbf{User:} Please perform the Breadth-first Search algorithm for reachability. Edgelist: [(0, 1), (0, 2), (0, 3), (1, 3)]; Source Node: 3. List all reachable nodes ordered lexicographically from smallest to largest.\\[0.75em]
        \textbf{Assistant:} [0, 1, 2, 3]
    \end{tcolorbox}

    \begin{tcolorbox}[colback=white!10!white,colframe=olive!75!black,title= Intermediate Steps, sharp corners, box align=top, width=\linewidth] % Ensure full width
        \begin{tikzpicture}[scale=0.7] % Adjust the scale here
            \node[circle, draw] (0) at (0,0) { 0};
            \node[circle, draw] (1) at (1.5,1.5) { 1};
            \node[circle, draw] (2) at (3,0) { 2};
            \node[circle, draw] (3) at (-1.5,1.5) { 3};
            \draw (0) -- node[above] {} (1);
            \draw (0) -- node[above] {} (2);
            \draw (0) -- node[left] {} (3);
            \draw (1) -- node[below] {} (3);
        \end{tikzpicture}
        \\[1em]
        \textbf{User:} Please perform the Breadth-first Search algorithm for reachability. Edgelist: [(0, 1), (0, 2), (0, 3), (1, 3)]; List all known reachable nodes ordered lexicographically from smallest to largest.\\[0.75em]
        \textbf{Assistant:} [0, 1] 
        \\
        • \\
        • \\
        • \\[0.75em]
        \textbf{User:} List all known reachable nodes ordered lexicographically from smallest to largest. \\[0.75em]
        \textbf{Assistant:} [0, 1, 2, 3]
    \end{tcolorbox}

    \begin{tcolorbox}[colback=white!10!white,colframe=teal!75!black,title= Intermediate Steps with Hints, sharp corners, box align=top, width=\linewidth] % Set width to full text width
        \begin{tikzpicture}[scale=0.7] % Adjust the scale here
            \node[circle, draw] (0) at (0,0) { 0};
            \node[circle, draw] (1) at (1.5,1.5) { 1};
            \node[circle, draw] (2) at (3,0) { 2};
            \node[circle, draw] (3) at (-1.5,1.5) { 3};
            \draw (0) -- node[above] {} (1);
            \draw (0) -- node[above] {} (2);
            \draw (0) -- node[left] {} (3);
            \draw (1) -- node[below] {} (3);
        \end{tikzpicture}
        \\[1em]
        \textbf{User:} Please perform the Breadth-first Search algorithm for reachability. Edgelist: [(0, 1), (0, 2), (0, 3), (1, 3)]; Source Node: 3. Queue: [3], Dequeue: 3, Unvisited neighborhood of 3: [0, 1]. List all known reachable nodes ordered lexicographically from smallest to largest. \\[0.75em] 
        \textbf{Assistant:} [0,1] 
        \\[0.5em]
        • \\
        • \\
        • \\[0.5em]
        \textbf{User:} Queue: [2], Dequeue: 2, Unvisited neighborhood of 2: []. List all known reachable nodes ordered lexicographically from smallest to largest. \\[0.75em]
        \textbf{Assistant:} [0, 1, 2, 3] 
    \end{tcolorbox}
} 
    \caption{Examples of the Input-Output, Intermediate Steps, and Intermediate Steps with Hints prompting strategies.}
    \label{prompting_strategies}
\end{figure}

\begin{figure}[h]
    \centering  
    \input{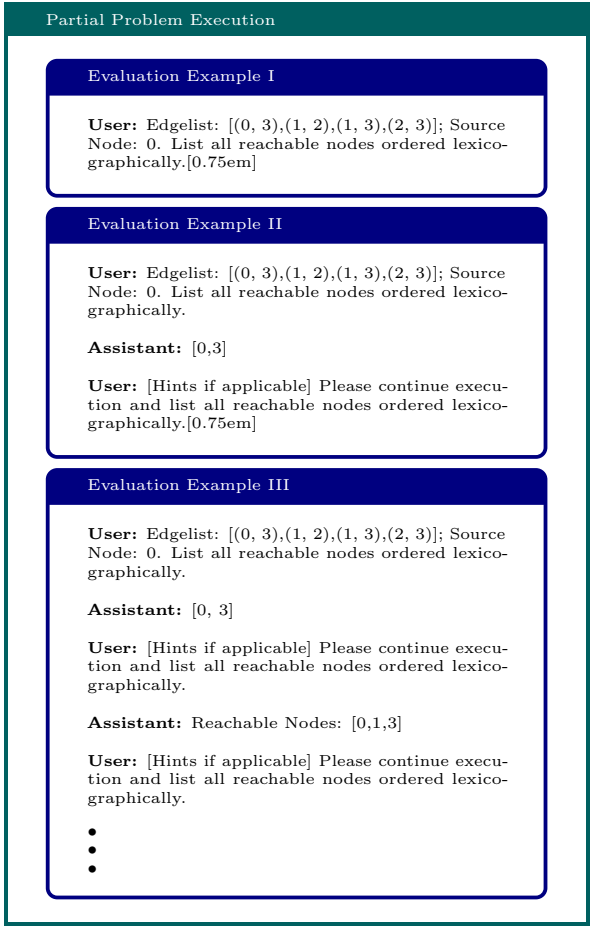} % Adjust the path as needed
    \caption{Illustration of the construction of partial algorithm executions evaluation.}
    \label{evaluation_examples}
\end{figure}

We construct the dataset for the \benchmark~with the goal of using a diverse subset of the graph algorithms covered in the CLRS benchmark to stress test LLMs on multistep graph reasoning problems. 
We provide three prompt reasoning strategies for fine-tuned models in the  \benchmark~(see Figure \ref{prompting_strategies}):
\begin{itemize}
    \item \textbf{Input-Output} (IO): Trajectory consists of one prompt, which presents the problem statement and the query for the final step solution \cite{besta_topologies_2024}.
    \item \textbf{Intermediate Steps} (IS): Trajectory consists of one or more prompts, of which the first presents the problem statement and queries for the next intermediate solution, which may be the final step output.
    \item \textbf{Intermediate Steps with Hints} (ISH): Similar to IS, however, intermediate prompts contain selected CLRS hints in addition to the intermediate solution query.
\end{itemize}
Following prior works, the Intermediate Steps and Intermediate Steps with Hints formats use teacher forcing while fine-tuning \cite{velickovic2021neural, velickovic2022clrs}. The evaluation task is structured as a series of chats in which the model is prompted to complete the \textit{next step} in the algorithmic execution with the correct prior steps (if any) included in the context window, as shown in Figure \ref{evaluation_examples}.

\section{Experimental Settings}
\subsection{Foundation Models}
We use Llama-3-8B, Llama-3-Instruct-8B, Mistral-7B-v0.3, Mistral-7B-Instruct-v0.3, and GPT-4o to establish baseline algorithmic reasoning performance on the \benchmark. These models were chosen due to their state-of-the-art (SOTA) capabilities in natural language processing and reasoning tasks \citep{openai2023gpt4, touvron2023llama, jiang2023mistral7b}. Due to cost and resource constraints, each of the Llama-3 and Mistral-v0.3 accuracies represents the average of 2 runs, and each GPT-4 accuracy is the result of 1 run.

\subsection{Data splits} 
Given that the number of training examples is limited only by the number of possible graph orientations of a given size, we used preliminary experiments to determine the minimum number of full\footnote{Each trajectory is divided into partial examples, as shown in the prior section. This results in roughly 3,000 total training examples.} trajectories (1000) for training loss to converge after 1 epoch in fine-tuned models across all algorithms. We divided the data into the traditional 80:10:10 split, which results in a split of 1000:125:125 trajectories for graph sizes greater than 6 and 800:112:112 for graphs of size 5. 

\begin{figure*}[ht!]
  \centering
\includegraphics[scale=0.32]{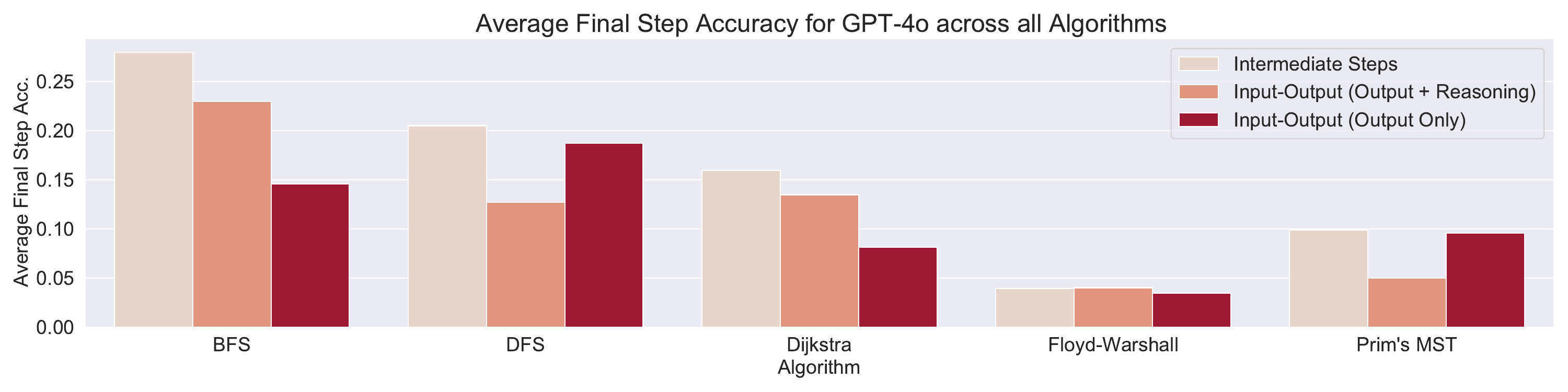}
  \caption{Average accuracy of GPT-4o models across all graph sizes. Output Only models were instructed to only provide the answer, while Output + Reasoning methods were permitted to use intermediate reasoning.}
\label{avg_gpt_algorithm_performance}
\end{figure*}

\begin{figure*}[ht!]
  \centering
\includegraphics[scale=0.32]{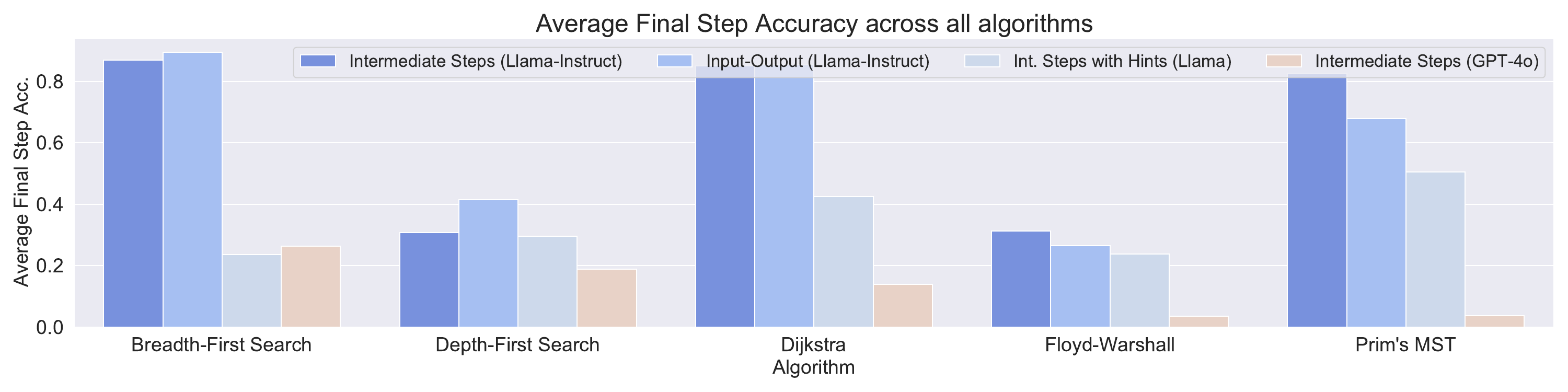}
  \caption{Average Final Step Accuracy of the best performing models.}
\label{avg_fs_algorithm_performance}
\end{figure*}

\begin{figure*}[ht!]
  \centering
\includegraphics[scale=0.32]{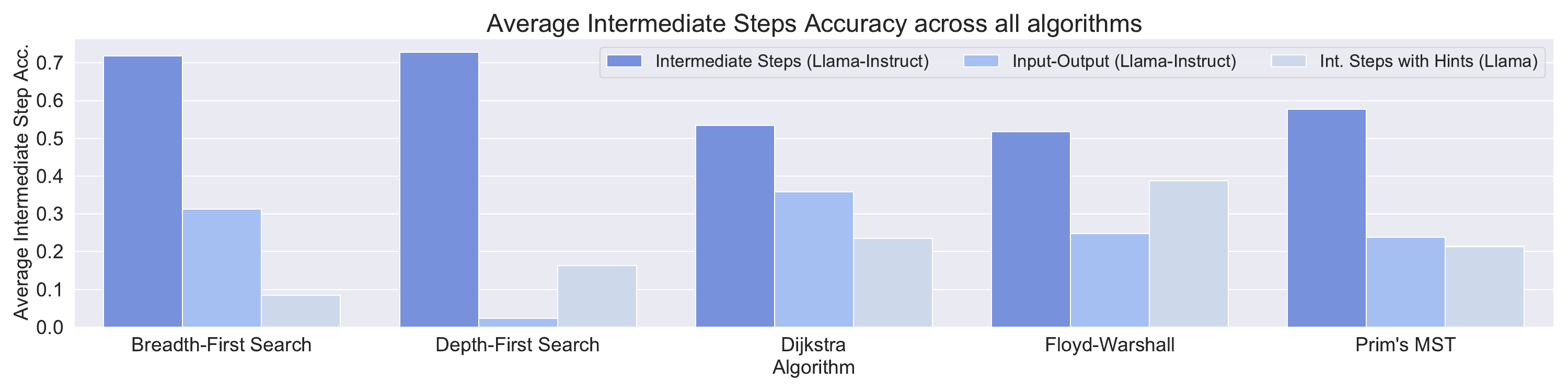}
  \caption{Average Intermediate Step Accuracy of the best performing models.}
\label{avg_is_algorithm_performance}
\end{figure*}

\begin{figure*}[ht!]
  \centering
\includegraphics[scale=0.32]{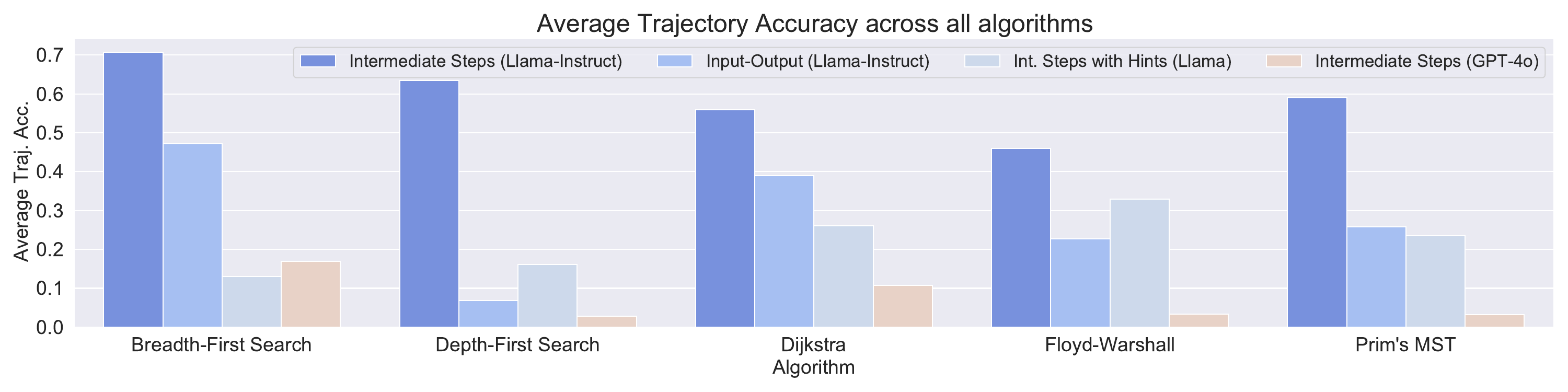}
  \caption{Average Trajectory Accuracy of the best performing models.}
\label{avg_traj_algorithm_performance}
\end{figure*}

\subsection{Evaluation Metrics} 
We use exact-match accuracy as our primary evaluation metric, following prior works \cite{wang_can_nodate, ge_graph_2024}.\footnote{We provide evaluations of the model using metrics, giving partial credit in the appendix.} 
We provide the following exact-match-accuracy-based evaluations of the baseline models:
\begin{itemize}
    \item \textbf{Final Step Accuracy}: Accuracy of the final solution step of an algorithmic trajectory. This metric provides an evaluation setting comparable to that of prior works. 
    \item \textbf{Intermediate Steps Accuracy}: Accuracy of all steps in the algorithmic trajectory \textit{except the final solution state}.
    \item \textbf{Trajectory Accuracy}: Accuracy of all steps in the algorithmic trajectory (weighted equally).
\end{itemize}
Please note that the experimental results we provide are comprised of the \textit{averages} of each of the above metric across evaluation examples for a given baseline setting.

\section{Results}

\subsubsection{Intermediate steps significantly improve multistep reasoning in fine-tuned models.}
Our results show that ``guiding'' the zero-shot LLM (GPT-4o) through explicit intermediate steps provides significant performance improvements. We also observe that the fine-tuned IS models outperform IO models on Avg. Final Step Accuracy on most algorithms, as shown in  Figure~\ref{avg_fs_algorithm_performance}. 
  
Furthermore, we find a statistically significant relationship between the Final Step Accuracy and the Intermediate Step Accuracy for many models trained with intermediate steps. We also observe significantly higher performance on the IS models when considering the accuracy across the whole problem trajectory, as shown in Figure \ref{avg_traj_algorithm_performance}. 
This indicates that for problems where correctness across multiple steps is required, fine-tuning on intermediate steps provides significant benefits to overall accuracy. 
These results support our assertion that incorporating intermediate steps into the training and evaluation pipeline for both fine-tuned models and in-context learning for zero-shot models confers performance benefits on graph algorithmic reasoning tasks. 

\subsubsection{Fine-tuned LLMs outperform the current state-of-the-art zero-shot model on graph algorithmic reasoning tasks.}
Our results indicate that fine-tuning even relatively small language models for complex graph reasoning tasks provides vast performance enhancements over the state-of-the-art zero-shot model (GPT-4o), as shown in both Figure~\ref{avg_fs_algorithm_performance}. This supports the findings of prior work in comparing small, fine-tuned LLMs to large-scale LLMs on multistep mathematical reasoning problems and demonstrates that they apply to multistep graph reasoning problems \cite{fu2023specializingsmallerlanguagemodels}.

\subsubsection{Models fine-tuned to leverage intermediate steps are sensitive to additional information} 
We observe that the inclusion of additional context for intermediate steps (translated from CLRS hints) hinders performance, as shown in Figure \ref{avg_is_algorithm_performance}\footnote{We provide examples of the hints used in the appendix.}. This is supported by prior work, and indicates that while the models benefit from the intermediate step structure, further work is necessary to determine how to incorporate intermediate step context \cite{BevilacquaNeuralAlg}. 

\begin{figure*}[ht!]
\centering
\begin{subfigure}{\textwidth}
  \centering
  \includegraphics[width=\textwidth]{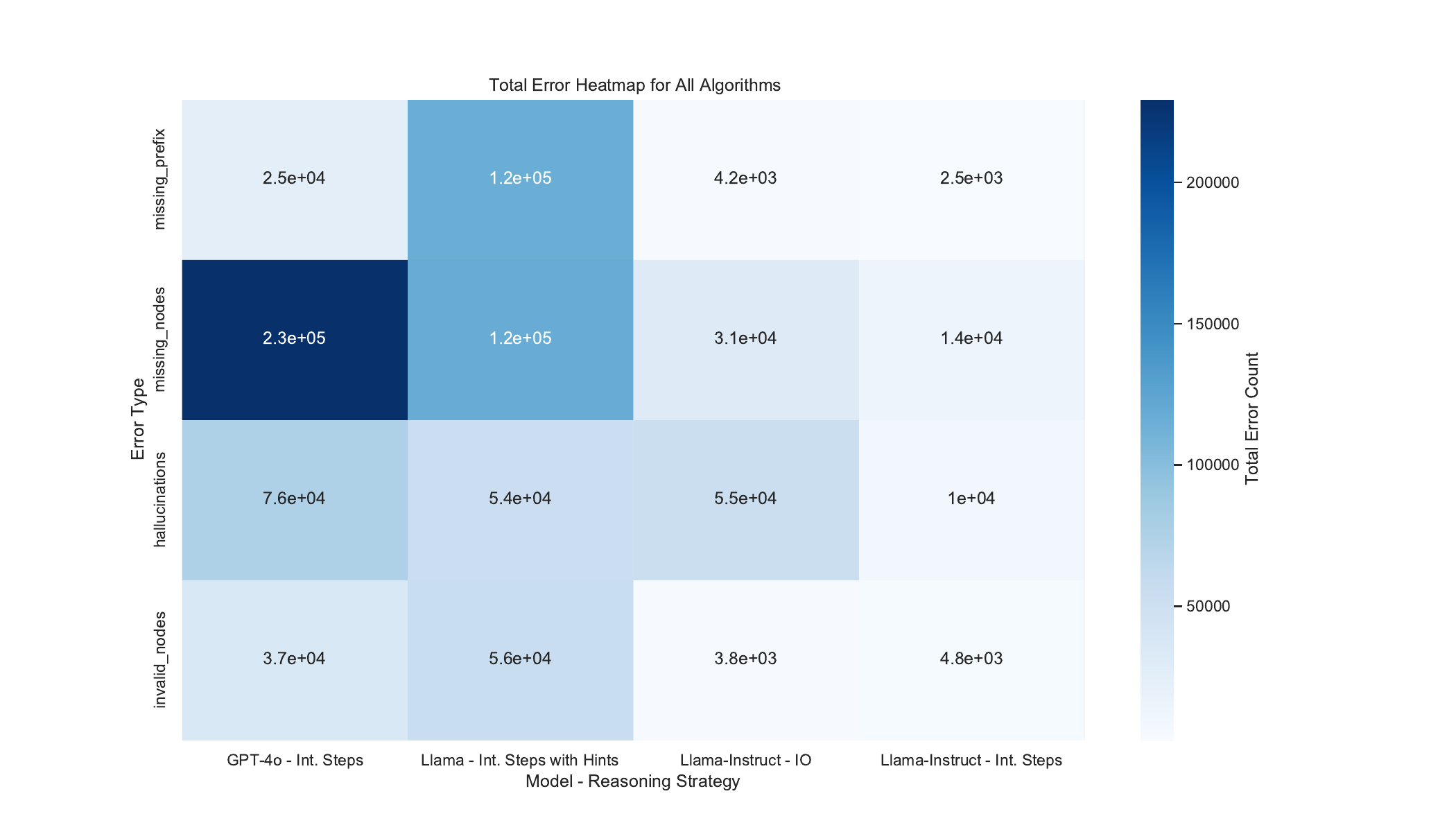}
  \caption{Heatmap of Error Types}
  \label{error_heatmap}
\end{subfigure}

\vspace{0.3cm} % Adds space between the first and second set of images

% Row 1: Three images
\begin{subfigure}{0.32\textwidth}
    \centering
    \includegraphics[width=\textwidth]{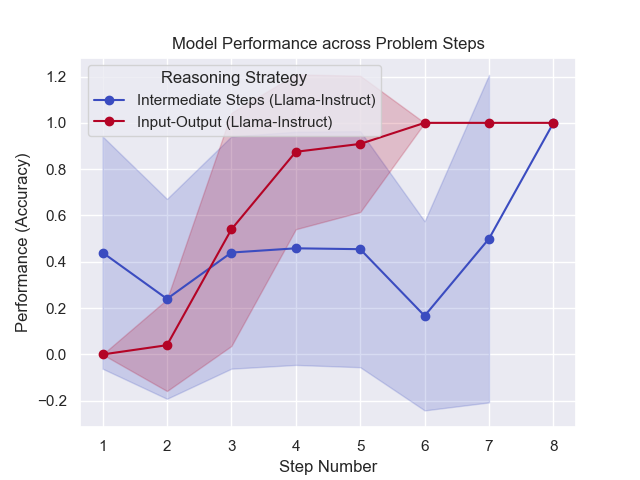}
    \caption{Breadth-First Search}
    \label{fig:ra_alg1}
\end{subfigure}
\hfill
\begin{subfigure}{0.32\textwidth}
    \centering
    \includegraphics[width=\textwidth]{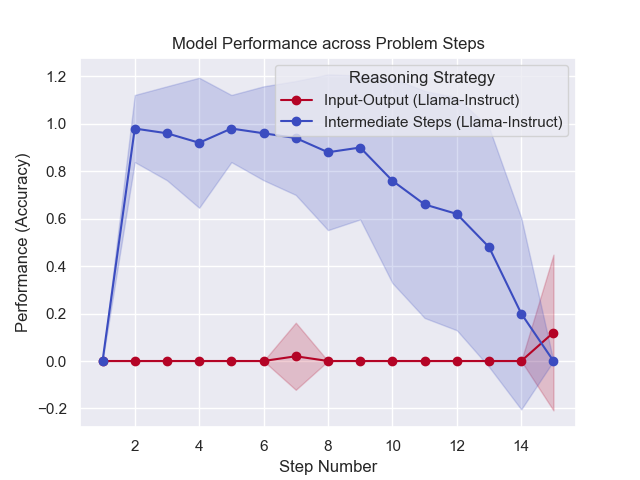}
    \caption{Depth-First Search}
    \label{fig:ra_alg2}
\end{subfigure}
\hfill
\begin{subfigure}{0.32\textwidth}
    \centering
    \includegraphics[width=\textwidth]{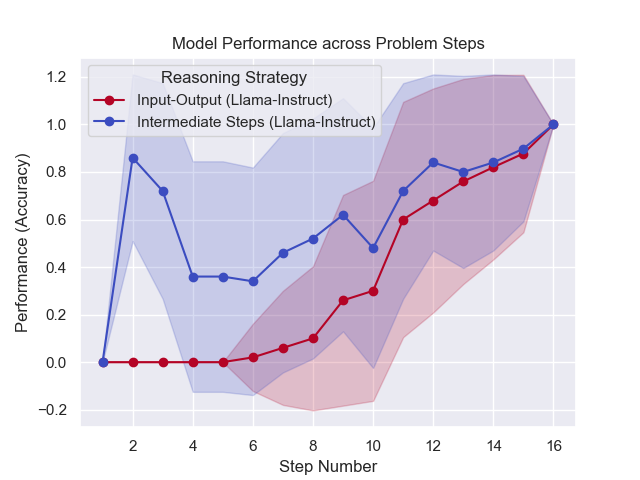}
    \caption{Dijkstra}
    \label{fig:ra_alg3}
\end{subfigure}

% Row 2: Two images, centered with \hfill to adjust for the gap
\begin{subfigure}{0.32\textwidth}
    \centering
    \includegraphics[width=\textwidth]{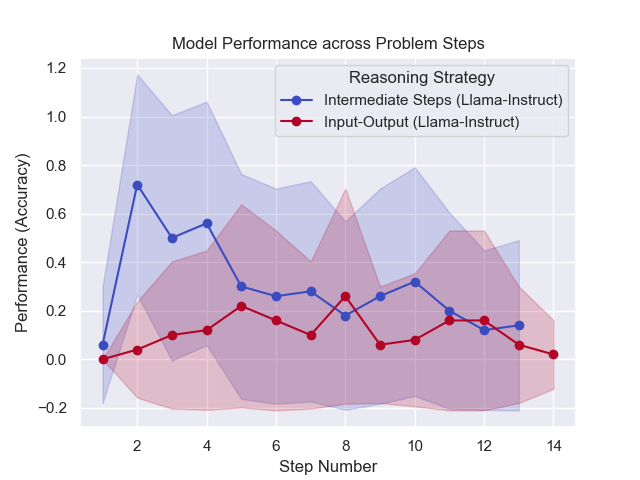}
    \caption{Floyd-Warshall}
    \label{fig:alg4}
\end{subfigure}
\begin{subfigure}{0.32\textwidth}
    \centering
    \includegraphics[width=\textwidth]{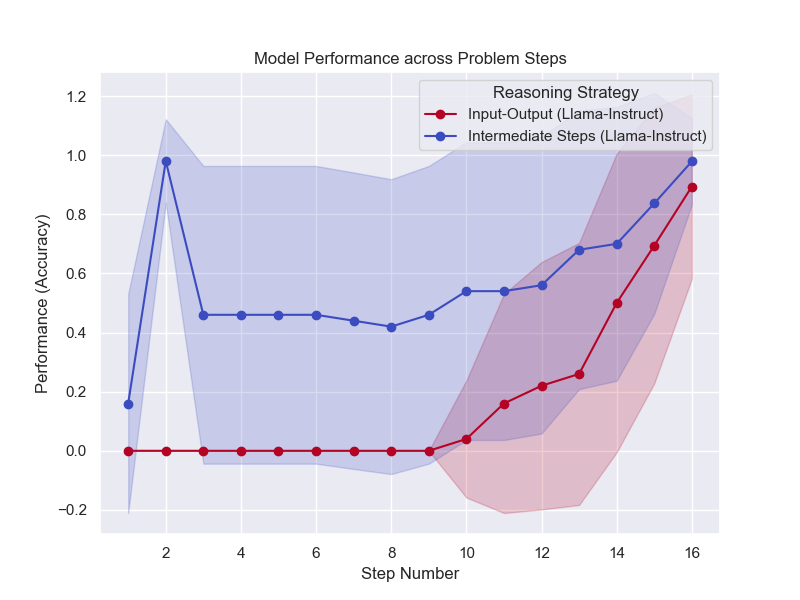}
    \caption{Prim's MST}
    \label{fig:alg5}
\end{subfigure}
\hfill
\caption{Error analysis and algorithm-specific trajectory analyses. Sub-figure (a) shows a heatmap of the errors we track and subfigures (b)-(f) show the average performance at step \textit{i} for each algorithm. Please note that these visualizations cover graph sizes 5-15 and see Appendix for full.}
\label{fig:ra_comprehensive}
\end{figure*}

\section{Results Analysis}
We provide a detailed view of the average performance of the IO and IS models on problem steps, as well as the types of errors commonly observed in the predictions of our models in Table \ref{error_heatmap}. 

\subsection{Error Analysis}
To better understand the challenges our models encounter, we categorize and examine commonly errors observed during prediction, as summarized below.
\begin{enumerate}
    \item \textbf{Missing Prefix.} Model prediction does not include the algorithm-specific prefix that denotes the start of the answer (See Appendix for examples of the expected format). We count only the first mistake in each prediction in our analysis. 
    \item \textbf{False Negatives.} The model prediction lacks an expected node or edge in the answer. We count each individual miss in our analysis.
    \item \textbf{Hallucinations.} The model predicts the false inclusion of nodes or edges in the answer. We count each individual addition in our analysis.
    \item \textbf{Invalid Answer Items.} Model prediction includes answer items in an incorrect format. For example, if we expect a list of reachable node indices (\([1, 2, \ldots]\)), but we get a list of text (\([ \text{node}_1, \text{node}_2, \ldots ]\)). We count only the first mistake in each prediction in our analysis. 
\end{enumerate}
We observe that the number of errors for a given model-prompting strategy follows the expected positive correlation between the number of errors of each type and the average performance of an algorithm (vertical axis). 
We also observe that models are prone to Hallucinations, False Negatives, or Missing Prefixes than they are to Invalid Answer Items.
We note that missing prefixes pose a consistent challenge for all models. 
In addition, we observe that weaker models are more likely to exclude nodes or edges (False Negatives) from answers than they are to add extra or nonexistent information (Hallucinations). We observe that the IO model is more likely to hallucinate additional nodes than to miss nodes, and also more likely to make either mistake than the IS model.
These trends indicate that stronger models are able to tell which nodes are likely to belong to an answer. Furthermore, it indicates that fine-tuning with intermediate steps allows models to gain a better understanding of which nodes to \textit{not} use. 

\subsection{Trajectory Analysis}
In Figure \ref{fig:ra_comprehensive}, we show the average accuracy for each step across algorithmic trajectories for graphs of size 5 through 15\footnote{We exclude graph sizes of 20 and 50 from this visualization for the sake of interpretability.}. We show the full average model performance on algorithmic executions in Figure \ref{fig:main}.

We show, in Figures \ref{fig:ra_alg1} and \ref{fig:alg2}, that on simpler problem types (Breadth-First Search and Depth-First Search), the trajectory lengths are shorter and fine-tuning on intermediate reasoning steps is less beneficial.
We also show that for more complex algorithms (Dijkstra, Floyd-Warshall, and Prim's MST), the trajectories are longer and intermediate steps provide more of a benefit.
We also observe that fine-tuned IS models vastly outperform fine-tuned IO models on intermediate step accuracy, as shown in Figure~\ref{avg_is_algorithm_performance}\footnote{The tables used to generate these figures are provided in the appendix.}.

\subsubsection{Breadth-First Search}
We first consider the errors committed by the IO and IS models on BFS. As shown in Figure \ref{error_heatmap}, both models commit relatively few errors. Furthermore, we see in Figure \ref{fig:ra_alg1} that the IO model performs well on the final steps of the problem across all graph sizes, while the IS model displays less consistent behavior. This result is intuitive considering that BFS trajectories are shorter than the other algorithms and the problem formulation requires only a list of node identifiers. This indicates that BFS is a simple enough problem for the IO model to grasp the intermediate steps without needing instruction. The IS model performance on intermediate steps is improved by fine-tuning, but final step accuracy is not improved and performance is less consistent with larger graph sizes.

\subsubsection{Depth-First Search}
However, both the IO and IS models commit a relatively high number of invalid format errors on DFS. This indicates that the models are sensitive to the format of the connected components representation used for this DFS problem. Additionally, we see a gradual decline in the accuracy of each subsequent step for the IS model; this indicates that the IS model is better on intermediate steps and performs better over shorter problem trajectories. On the other hand, we see that the IO model tends to only perform well on the final problem steps; Figure \ref{fig:scatterplots_all_algorithms} also shows that IO model performances degrade as graph size increases. These results suggest that the intermediate step fine-tuning vastly improves intermediate step accuracy at the cost of some final step accuracy.

\subsubsection{Floyd-Warshall}
Model performances on Floyd-Warshall performance drop as trajectory length increases. This result follows naturally when considering both the long lengths of problem trajectories and large solution sizes of Floyd-Warshall problems. This difficulty is reflected in the higher number of False Negatives and Hallucinations committed by both IO and IS models. We also observe in Figure \ref{fig:scatterplots_all_algorithms} that IO models are relatively accurate on the final steps of smaller graphs, but performance collapses as graph sizes increase. On the other hand, we see that IS models are accurate on smaller graph sizes, and the trend in intermediate step and final step accuracies remain consistent across graph sizes. This indicates that the intermediate step fine-tuning makes the model more robust to larger problem sizes.

\subsubsection{Prim's MST \& Dijkstra}
Models trained on Prim's and Dijkstra yield similar results and consistently perform well on final step accuracy. This follows intuition, as both problems have similar trajectory lengths and similar solution sizes. Furthermore, we observe similar error trends across both algorithms with IO models tending to commit frequent hallucinations, which likely results from the IO model predicting its final step solution at intermediate steps. We also observe in Figure \ref{fig:scatterplots_all_algorithms} that IO models are relatively accurate across the problem trajectories of smaller graphs, but lose intermediate step performance and gain final step performance as graph sizes increase. On the other hand, we see that IS models display a dip in performance around the $8^{th}$ solution step, but improve on both intermediate and final step accuracy as graph sizes increase. This indicates that both the intermediate and final step accuracies of the IS models benefit from intermediate step fine-tuning.

We find that more complex problems (FLoyd-Warshall, Prim's MST, and Dijkstra) tend to require longer trajectories, and that IS models are more robust to longer trajectories and increasing graph sizes on average across the 5 algorithms than the IO models.
\section{Conclusion}

In this work, we present the \benchmark, a comprehensive benchmark for algorithmic reasoning on graph-structured data that utilizes the CLRS framework. We show that using intermediate steps both in in-context learning and fine-tuning improves the performance of SOTA foundation LLMs on five classical computer science graph algorithms. 
These findings highlight the potential of training LLMs to follow multistep reasoning processes for improving multistep algorithmic problem-solving on graphs.

It is our hope that this benchmark will improve access to algorithmic reasoning tasks in natural language, address the need for a diverse and comprehensive natural language multistep reasoning dataset.
% \section*{Ethics Statement}

 \section*{Limitations}
This evaluation is based on selected graph reasoning tasks, and we only evaluate a subset of available large language models. The observations and conclusions made in this study may need to scale up for a larger collection of language models to be considered as a general law of LLMs' graph reasoning capabilities. Furthermore, we note that our results may be subject to prompt sensitivity based on our query structure.

\section*{Acknowledgements}
We extend our heartfelt gratitude to Brandon Lo, Inman Costa, and Aubrey Wu for their contributions to the implementation of our methods. Additionally, we are thankful to Jeehyun Hwang for his expert advice and constructive feedback on our manuscript.
Lastly, we are deeply thankful to all those who contributed to the CLRS benchmark, without whom our work would not have been possible.

The contributions of everyone involved have been pivotal in the advancement of this research, and we are deeply appreciative of their efforts.

\bibliography{./bibliographies/custom}

\appendix
\section{Appendix}

\subsection{Frameworks} 
Our entire codebase is implemented in PyTorch. The implementations of the transformer-based models are extended from the Huggingface codebase.

\subsection{Hyperparameter Tuning and Model Settings}
All fine-tuning experiments were performed in half-precision using the aforementioned Huggingface codebase. The models for each algorithm were tuned for the optimal \textit{r} and \textit{alpha} values for parameter-efficient fine-tuning.

\subsection{Algorithmic Translation Details}

This section describes the translation of inputs, hints, and outputs for various algorithms used in our benchmarks. The details are derived from the CLRS benchmark and our own methodology.

\subsection{Algorithm Specifications}

Each algorithm specification includes the following fields:
\begin{itemize}
    \item \textbf{inputs}: The initial parameters or data structures required by the algorithm to execute.
    \item \textbf{hints}: Intermediate steps or states generated during the execution of the algorithm, useful for reasoning about the process.
    \item \textbf{outputs}: The final result produced by the algorithm.
\end{itemize}

\subsection{Translation of Inputs, Hints, and Outputs}

For each algorithm, we describe the translation steps from the raw data to the form used in our benchmarks.

\subsubsection{Breadth-First Search (BFS)}

\paragraph{Inputs:}
\begin{itemize}
    \item \textbf{Adjacency Matrix (adj)}: A matrix representing the connections between nodes in the graph.
    \item \textbf{Source Node (s)}: The starting node for the BFS algorithm.
\end{itemize}

\paragraph{Hints:}
\begin{itemize}
    \item \textbf{Reachable Nodes (reach\_h)}: Indicates which nodes are reachable from the source node.
    \item \textbf{Predecessor Nodes (pi\_h)}: Shows the predecessor of each node in the BFS tree.
\end{itemize}

\paragraph{Translation:}
\begin{itemize}
    \item \textbf{User Hint:} \texttt{Queue: \{queue\}, Dequeue: \{dequeue\_node\}, Unvisited neighborhood of \{dequeue\_node\}: \{neighborhood\}}
    \item \textbf{Assistant Hint:} \texttt{Reachable Nodes: \{reachable\_nodes\}}
\end{itemize}

\paragraph{Outputs:}
\begin{itemize}
    \item \textbf{Reachable Nodes List}: A list of nodes reachable from the source node.
\end{itemize}

\paragraph{Translation Steps:}
\begin{itemize}
    \item Convert the adjacency matrix to an edge list.
    \item Identify the source node.
    \item For hints, preprocess the hint matrices to extract reachable nodes and predecessors.
    \item Generate step-by-step hints detailing the queue state and reachable nodes.
\end{itemize}

\subsubsection{Depth-First Search (DFS)}

\paragraph{Inputs:}
\begin{itemize}
    \item \textbf{Adjacency Matrix (adj)}: A matrix representing the connections between nodes in the graph.
\end{itemize}

\paragraph{Hints:}
\begin{itemize}
    \item \textbf{Predecessor Nodes (pi\_h)}: Shows the predecessor of each node in the DFS tree.
    \item \textbf{Node Colors (color)}: Represents the state of each node (unvisited, visiting, visited).
    \item \textbf{Source Node (s)}: Indicates the source node.
\end{itemize}

\paragraph{Translation:}
\begin{itemize}
    \item \textbf{User Hint:} \texttt{Stack: \{stack\}, Pop Node: \{pop\_node\}, 1-hop Neighborhood of \{pop\_node\}: \{neighborhood\}}
    \item \textbf{Assistant Hint:} \texttt{Connected Components: \{components\}}
\end{itemize}

\paragraph{Outputs:}
\begin{itemize}
    \item \textbf{Connected Components}: A list of all connected components in the graph.
\end{itemize}

\paragraph{Translation Steps:}
\begin{itemize}
    \item Convert the adjacency matrix to an edge list.
    \item Identify the source node.
    \item For hints, preprocess the hint matrices to extract predecessors, node colors, and source node.
    \item Generate step-by-step hints detailing the stack state and connected components.
\end{itemize}

\subsubsection{Dijkstra's Algorithm}

\paragraph{Inputs:}
\begin{itemize}
    \item \textbf{Adjacency Matrix (adj)}: A matrix representing the connections between nodes in the graph.
    \item \textbf{Weights (A)}: A matrix representing the weights of the edges.
    \item \textbf{Source Node (s)}: The starting node for the algorithm.
\end{itemize}

\paragraph{Hints:}
\begin{itemize}
    \item \textbf{Distances (d)}: The current shortest distances from the source node to all other nodes.
    \item \textbf{Mark (mark)}: Indicates whether a node has been visited.
    \item \textbf{In Queue (in\_queue)}: Indicates whether a node is in the priority queue.
    \item \textbf{Current Node (u)}: The current node being processed.
\end{itemize}

\paragraph{Translation:}
\begin{itemize}
    \item \textbf{User Hint:} \texttt{Priority Queue: \{priority\_queue\}, Unvisited Nodes: \{unvisited\_nodes\}, Visited Nodes: \{visited\_nodes\}}
    \item \textbf{Assistant Hint:} \texttt{Distances: \{distances\}}
\end{itemize}

\paragraph{Outputs:}
\begin{itemize}
    \item \textbf{Distances}: The final shortest distances from the source node to all other nodes.
\end{itemize}

\paragraph{Translation Steps:}
\begin{itemize}
    \item Convert the adjacency matrix and weights to an edge list with weights.
    \item Identify the source node.
    \item For hints, preprocess the hint matrices to extract distances, marks, queue states, and current node.
    \item Generate step-by-step hints detailing the priority queue state, visited and unvisited nodes, and distances.
\end{itemize}

\subsubsection{Floyd-Warshall Algorithm}

\paragraph{Inputs:}
\begin{itemize}
    \item \textbf{Adjacency Matrix (adj)}: A matrix representing the connections between nodes in the graph.
    \item \textbf{Weights (A)}: A matrix representing the weights of the edges.
\end{itemize}

\paragraph{Hints:}
\begin{itemize}
    \item \textbf{Distance Matrix (D)}: The current shortest distances between all pairs of nodes.
\end{itemize}

\paragraph{Translation:}
\begin{itemize}
    \item \textbf{User Hint:} \texttt{Queue: \{queue\}, Dequeue \{dequeue\_node\}}
    \item \textbf{Assistant Hint:} \texttt{Distances: \{distances\}}
\end{itemize}

\paragraph{Outputs:}
\begin{itemize}
    \item \textbf{Distances}: The final shortest distances between all pairs of nodes.
\end{itemize}

\paragraph{Translation Steps:}
\begin{itemize}
    \item Convert the adjacency matrix and weights to an edge list with weights.
    \item For hints, preprocess the hint matrices to extract the distance matrix.
    \item Generate step-by-step hints detailing the distance matrix state and changes.
\end{itemize}

\subsubsection{Minimum Spanning Tree - Prim's Algorithm}

\paragraph{Inputs:}
\begin{itemize}
    \item \textbf{Adjacency Matrix (adj)}: A matrix representing the connections between nodes in the graph.
    \item \textbf{Weights (A)}: A matrix representing the weights of the edges.
    \item \textbf{Source Node (s)}: The starting node for the algorithm.
\end{itemize}

\paragraph{Hints:}
\begin{itemize}
    \item \textbf{Key (key)}: The current minimum weight to connect to the node.
    \item \textbf{Predecessor (pi\_h)}: The predecessor node in the MST.
    \item \textbf{Mark (mark)}: Indicates whether a node is included in the MST.
    \item \textbf{In Queue (in\_queue)}: Indicates whether a node is in the priority queue.
    \item \textbf{Current Node (u)}: The current node being processed.
\end{itemize}

\paragraph{Translation:}
\begin{itemize}
    \item \textbf{User Hint:} \texttt{Priority Queue: \{priority\_queue\}, Unvisited Nodes: \{unvisited\_nodes\}, Visited Nodes: \{visited\_nodes\}}
    \item \textbf{Assistant Hint:} \texttt{MST Edges: \{mst\_edges\}}
\end{itemize}

\paragraph{Outputs:}
\begin{itemize}
    \item \textbf{MST Edges}: The edges included in the minimum spanning tree.
\end{itemize}

\paragraph{Translation Steps:}
\begin{itemize}
    \item Convert the adjacency matrix and weights to an edge list with weights.
    \item Identify the source node.
    \item For hints, preprocess the hint matrices to extract key values, predecessors, marks, queue states, and current node.
    \item Generate step-by-step hints detailing the priority queue state, visited and unvisited nodes, and MST edges.
\end{itemize}

\subsubsection{Algorithm Translations}
Figure~\ref{fig:algorithm_translations} shows translations for algorithms.

\begin{figure}
    \centering
    \input{./Figures/Examples/app_examples}
    \caption{Algorithmic translations.}
    \label{fig:algorithm_translations}
\end{figure}

\subsection{Individual Prompt Strategy Performances Graphs}
\begin{figure*}[t!]
    \centering
    % First row of images
    \begin{subfigure}[b]{0.45\textwidth}
        \centering
        \includegraphics[width=\textwidth]{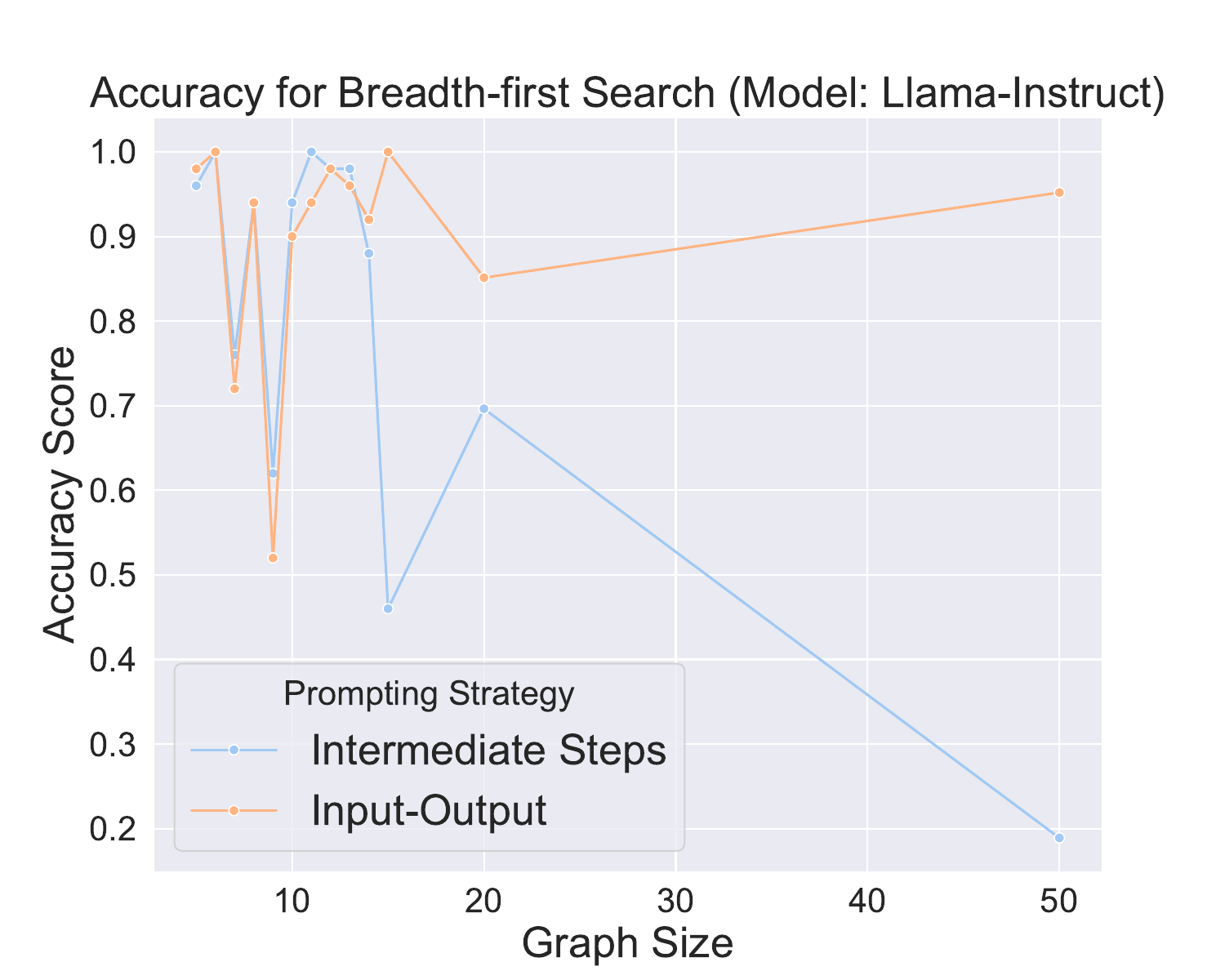}
        \caption{Breadth-First Search}
    \end{subfigure}
    \begin{subfigure}[b]{0.45\textwidth}
        \centering
        \includegraphics[width=\textwidth]{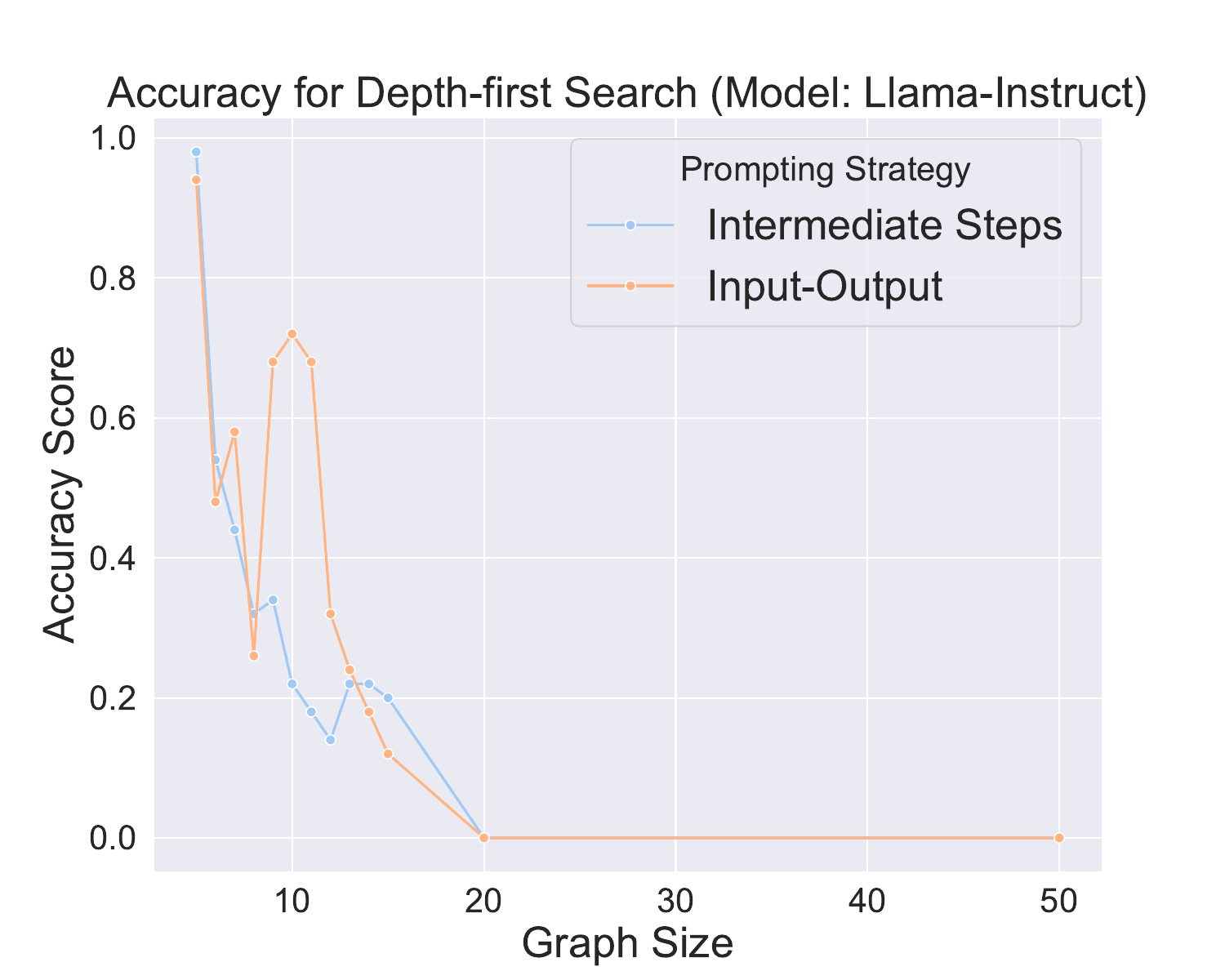}
        \caption{Depth-First Search}
    \end{subfigure}

    % Second row of images
    \begin{subfigure}[b]{0.45\textwidth}
        \centering
        \includegraphics[width=\textwidth]{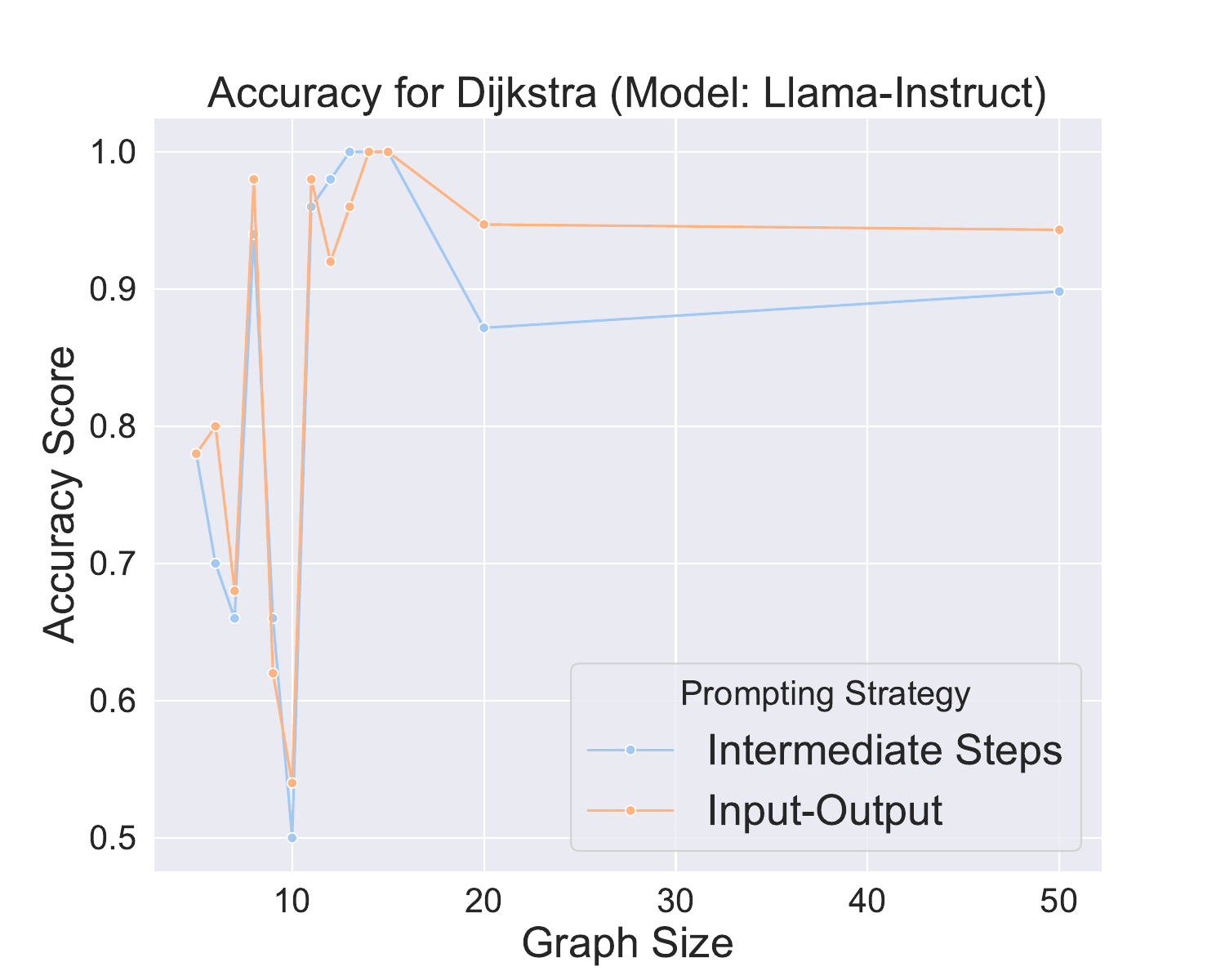}
        \caption{Dijkstra}
    \end{subfigure}
    \begin{subfigure}[b]{0.45\textwidth}
        \centering
        \includegraphics[width=\textwidth]{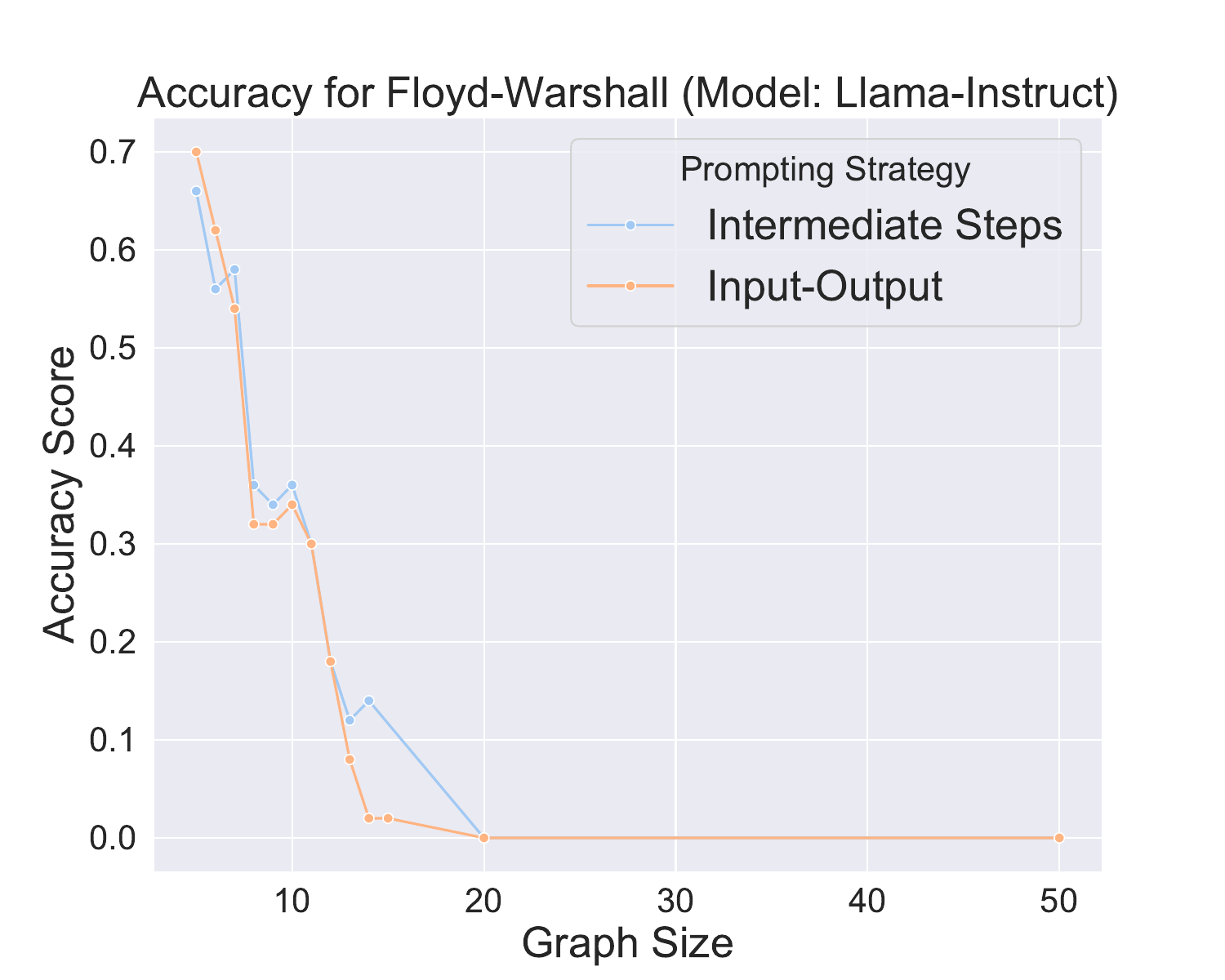}
        \caption{Floyd-Warshall}
    \end{subfigure}
   
    % Third row of images
    \begin{subfigure}[b]{0.45\textwidth}
        \centering
        \includegraphics[width=\textwidth]{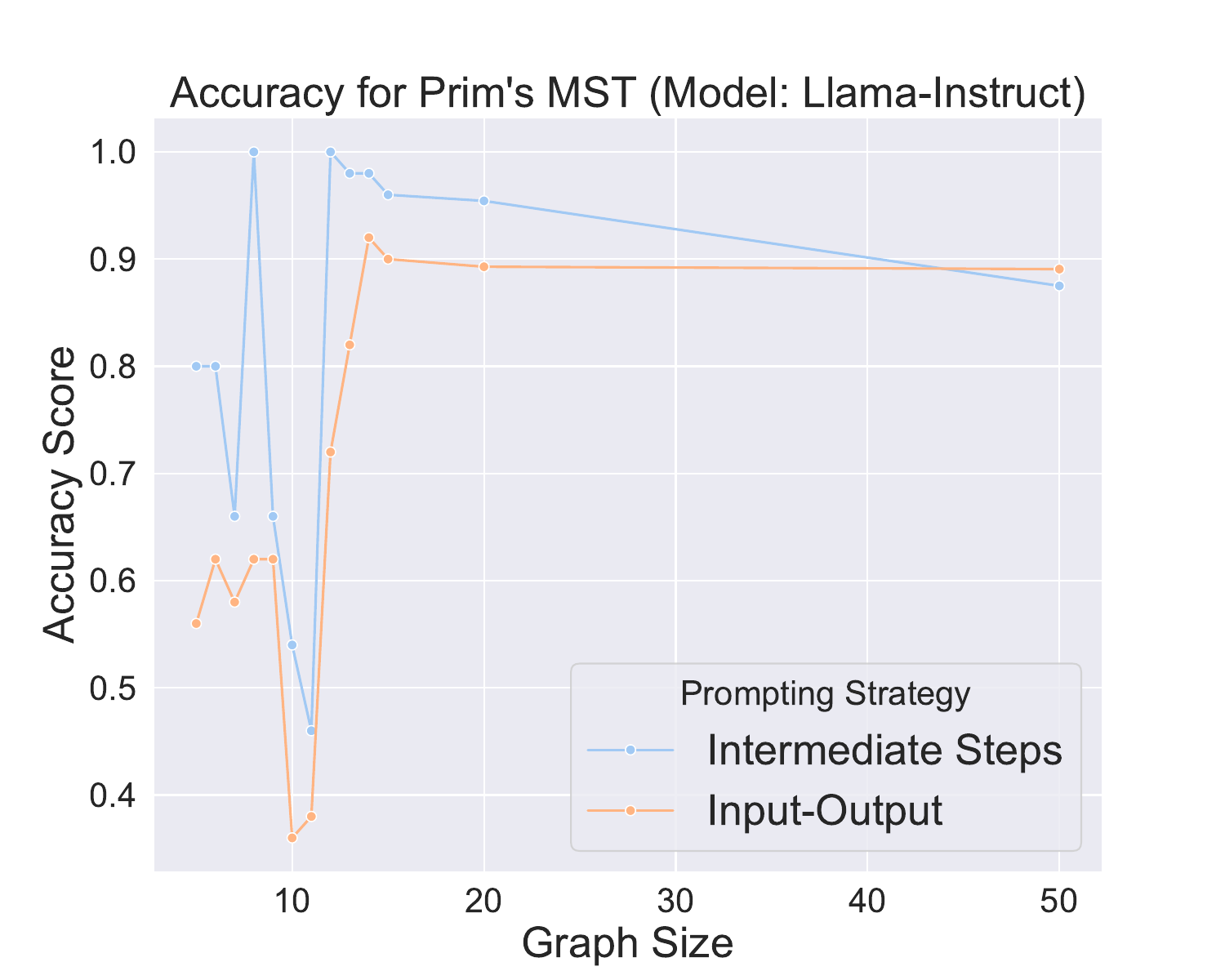}
        \caption{Prim's MST}
    \end{subfigure}
    
    \caption{Performance by graph size for BFS, DFS, Dijkstra, Floyd-Warshall, and Prim's MST algorithms.}
    \label{fig:main}
\end{figure*}

The performance graphs for individual prompt strategy is shown in Figure~\ref{fig:main} and Figure~\ref{fig:main2}.

\subsection{Hardware}
To run our experiments, we utilized a high-performance computing setup equipped with seven NVIDIA L40S GPUs and CUDA version 12.5.

\subsection{Error Heatmaps by Algorithm}
These figures are the algorithm-specific errors made by each of the best performing models.
\begin{figure*}[t!]
    \centering
    
    % Subfigure 1: Algorithm 1
    \begin{subfigure}{0.45\textwidth}
        \centering
        \includegraphics[width=\textwidth]{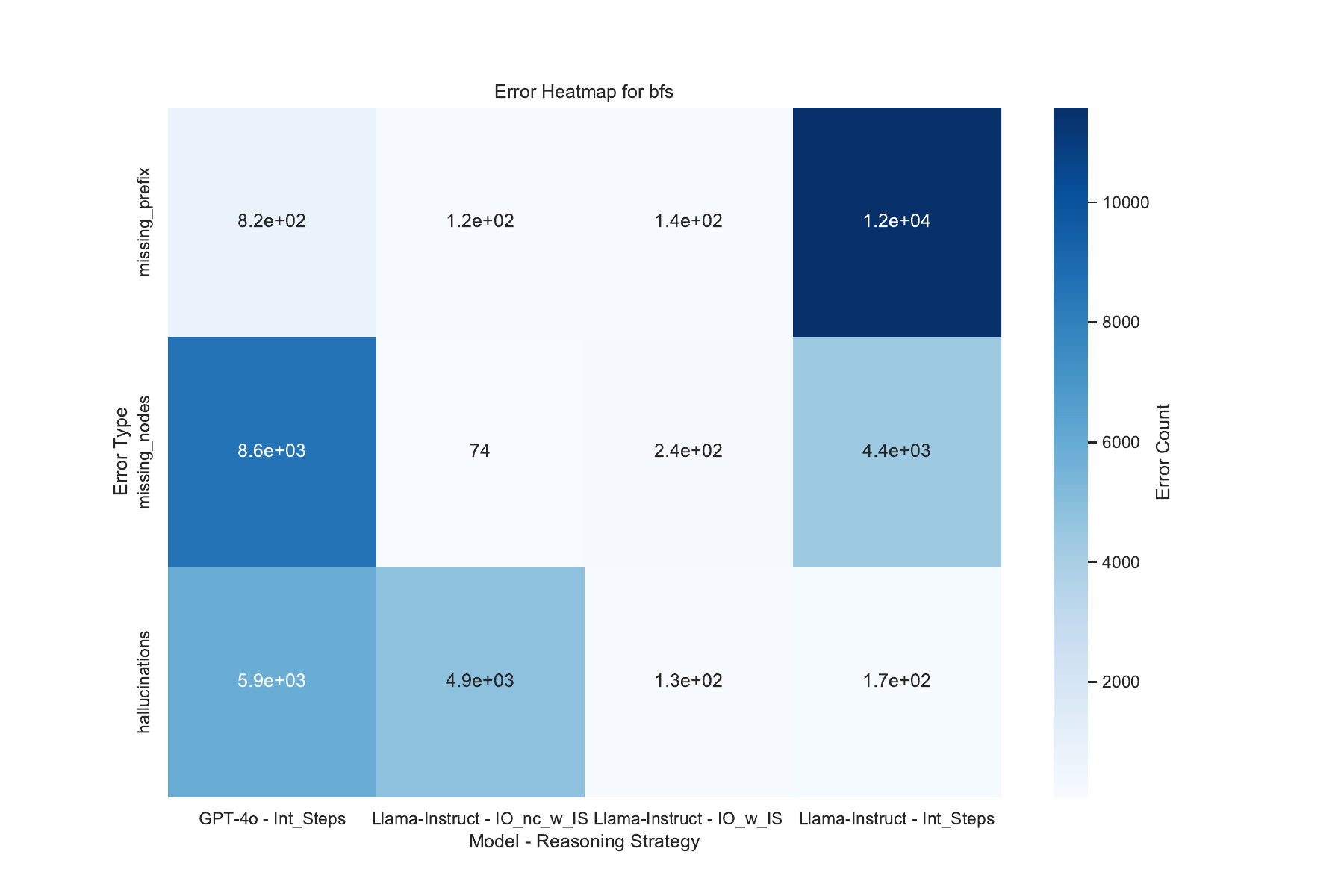}
        \caption{Breadth-First Search}
        \label{fig:alg1}
    \end{subfigure}
    \hfill
    % Subfigure 2: Algorithm 2
    \begin{subfigure}{0.45\textwidth}
        \centering
        \includegraphics[width=\textwidth]{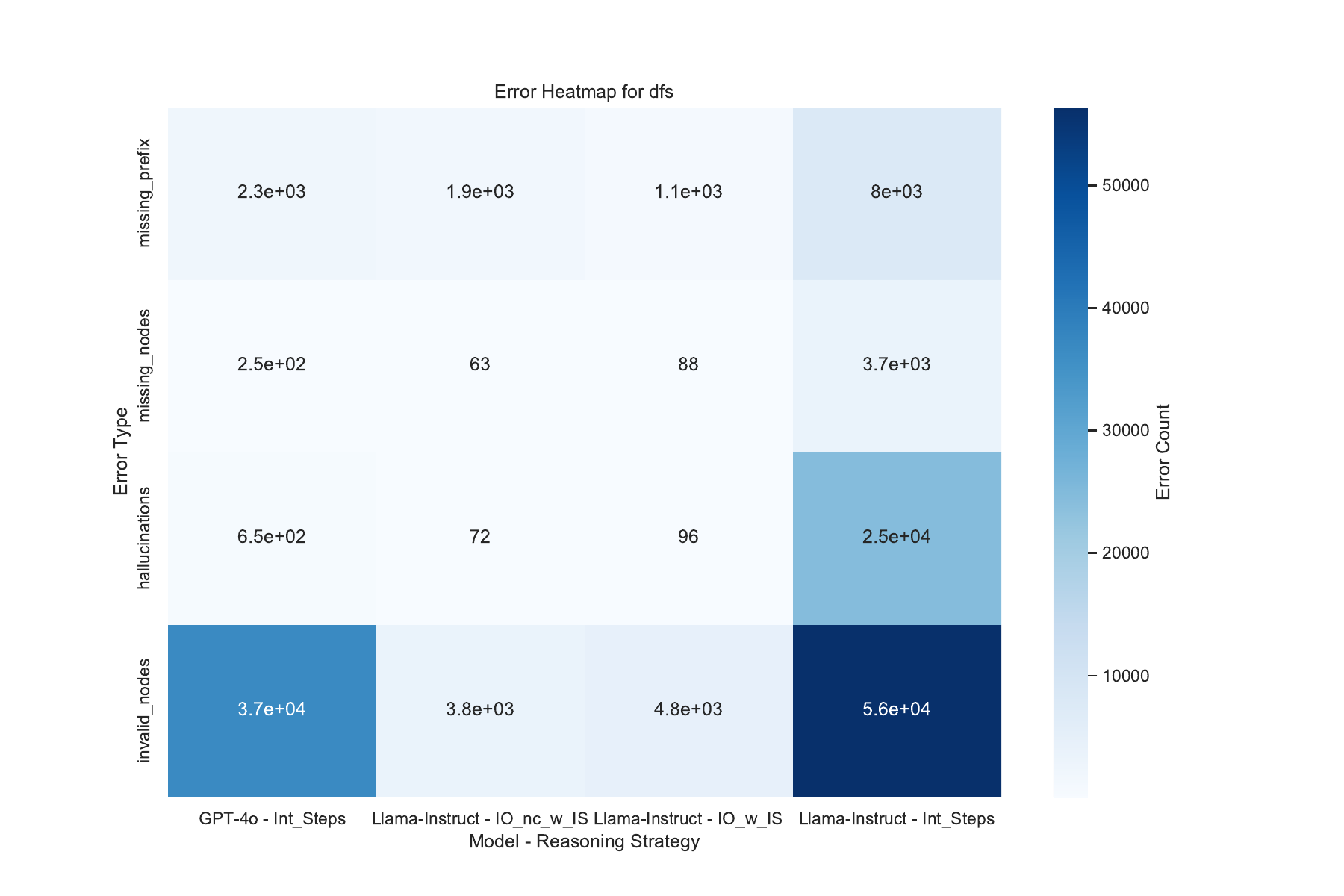}
        \caption{Depth-First Search}
        \label{fig:alg2}
    \end{subfigure}
    \hfill
    % Subfigure 3: Algorithm 3
    \begin{subfigure}{0.45\textwidth}
        \centering
        \includegraphics[width=\textwidth]{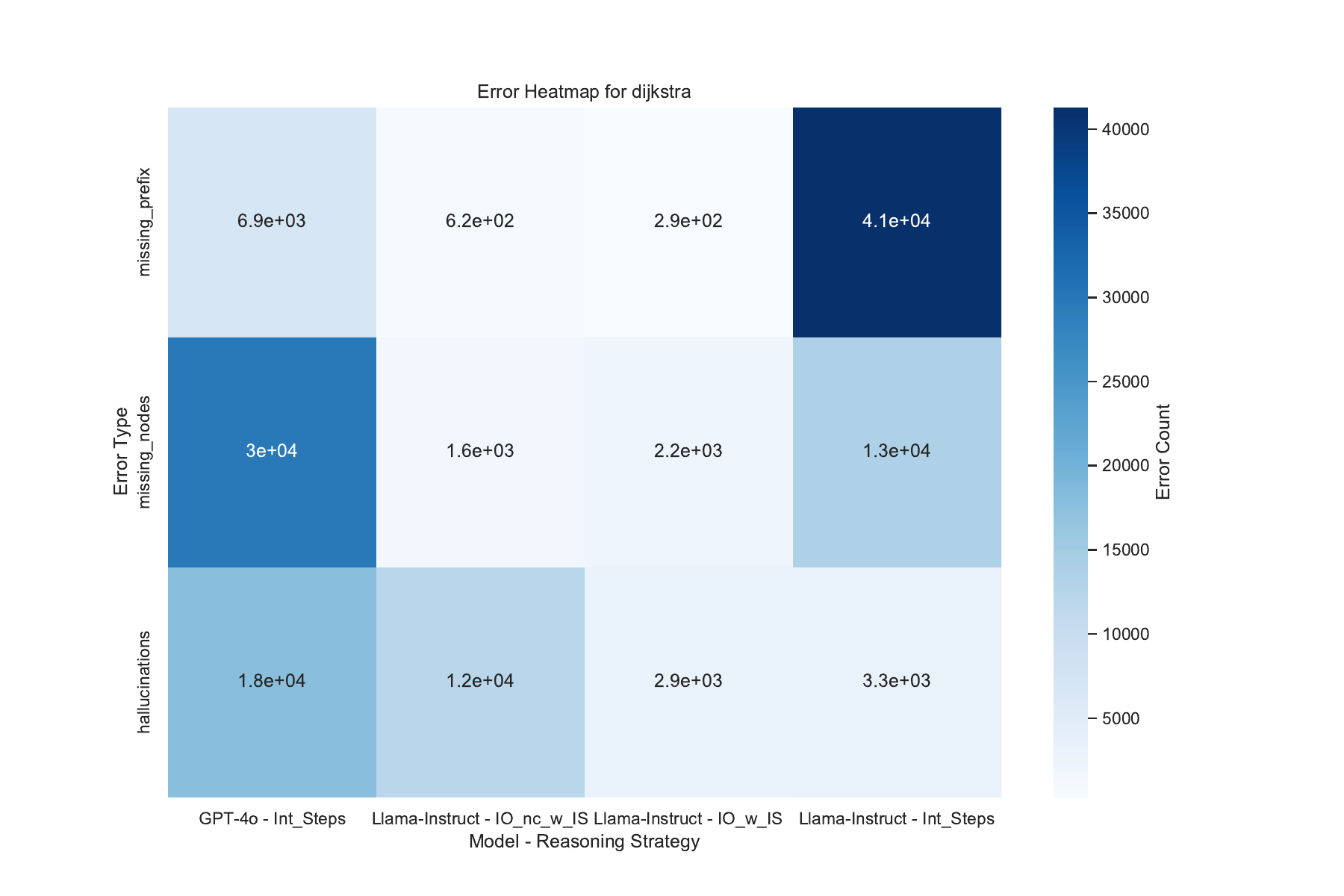}
        \caption{Dijkstra}
        \label{fig:alg3}
    \end{subfigure}
    
    % Subfigure 4: Algorithm 4
    \begin{subfigure}{0.45\textwidth}
        \centering
        \includegraphics[width=\textwidth]{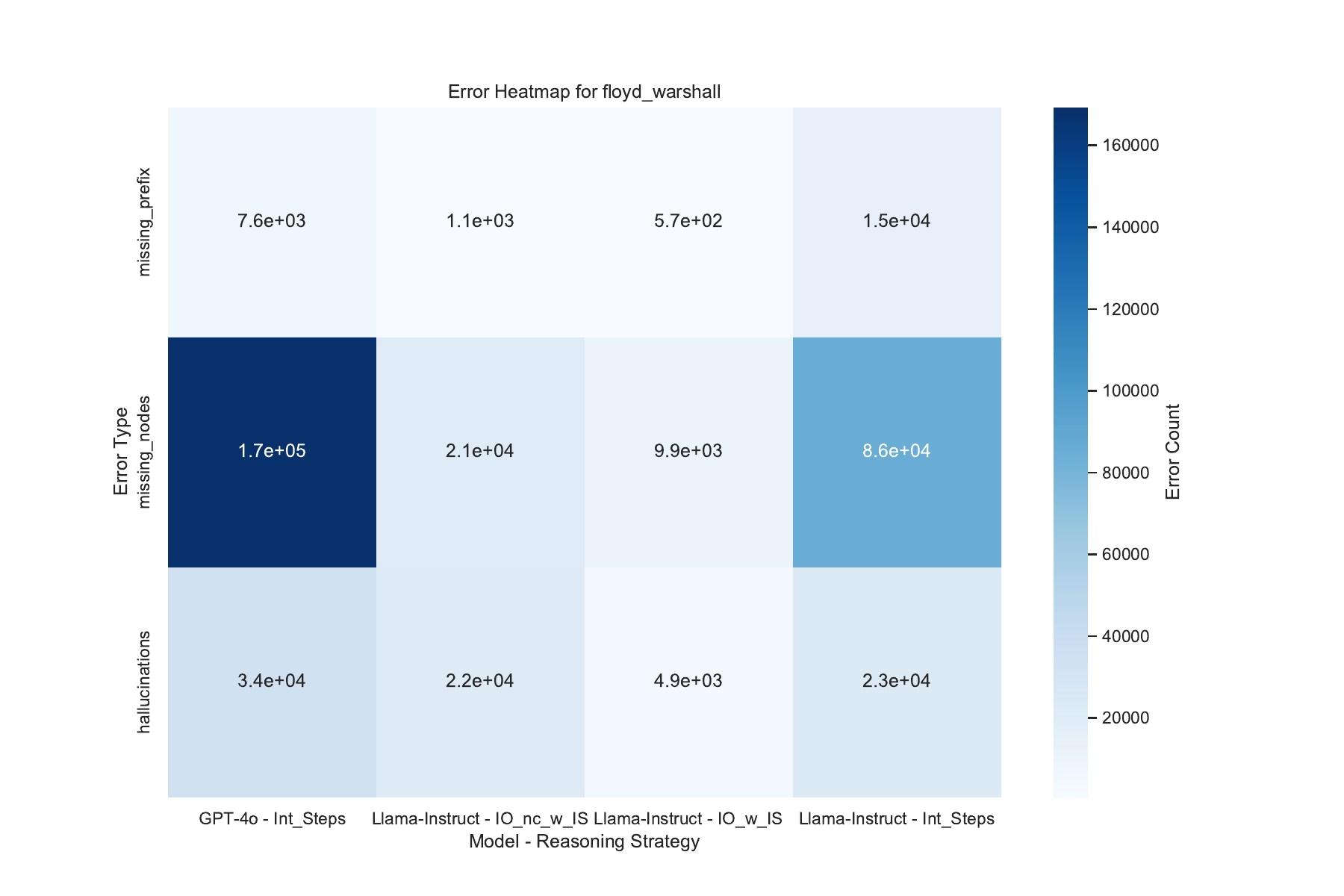}
        \caption{Floyd-Warshall}
        \label{fig:alg4}
    \end{subfigure}
    \hfill
    % Subfigure 5: Algorithm 5
    \begin{subfigure}{0.45\textwidth}
        \centering
        \includegraphics[width=\textwidth]{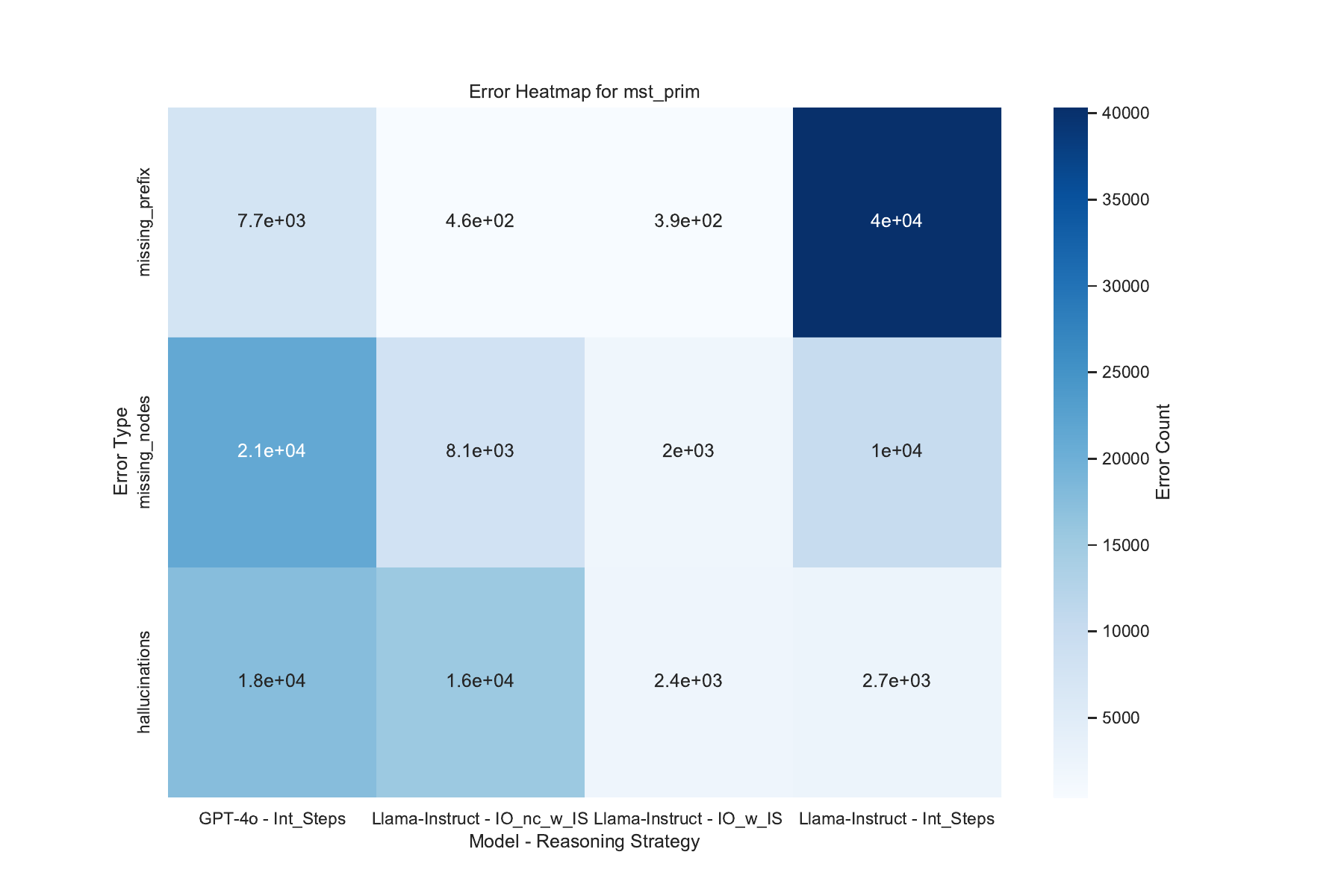}
        \caption{Prim's MST}
        \label{fig:alg5}
    \end{subfigure}

    \caption{Error heatmaps for each algorithm}
    \label{fig:heatmaps}
\end{figure*}

\subsection{Intermediate Steps Analysis}
These figures show the average accuracy for each step for each graph size by model. 
\begin{figure*}[ht!]
\centering
    % Row 1: BFS
    \begin{subfigure}{0.4\linewidth}
        \centering
        \includegraphics[width=\linewidth]{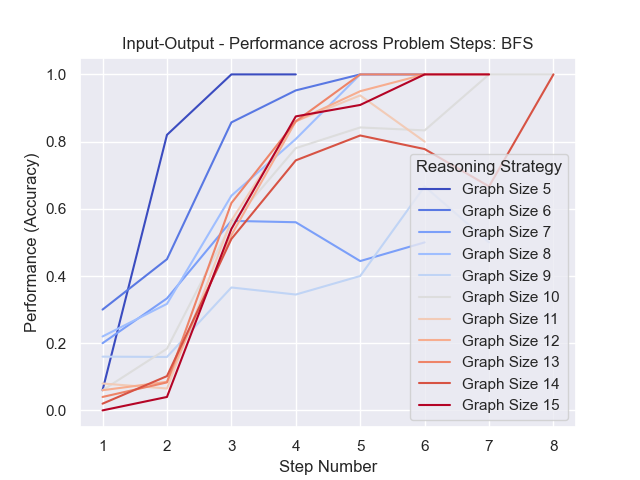}
        \caption{BFS - Input-Output}
    \end{subfigure}
    \hfill
    \begin{subfigure}{0.4\linewidth}
        \centering
        \includegraphics[width=\linewidth]{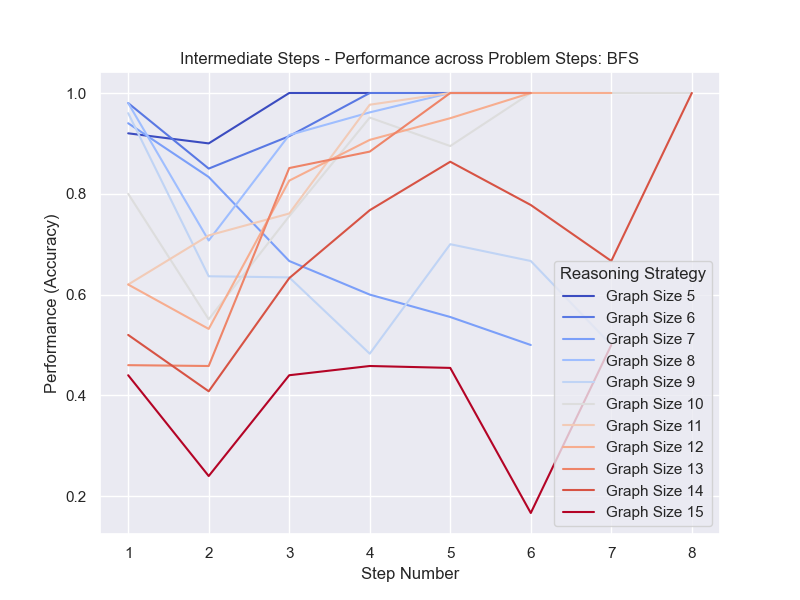}
        \caption{BFS - Intermediate Steps}
    \end{subfigure}
    
    % Row 2: DFS
    \begin{subfigure}{0.4\linewidth}
        \centering
        \includegraphics[width=\linewidth]{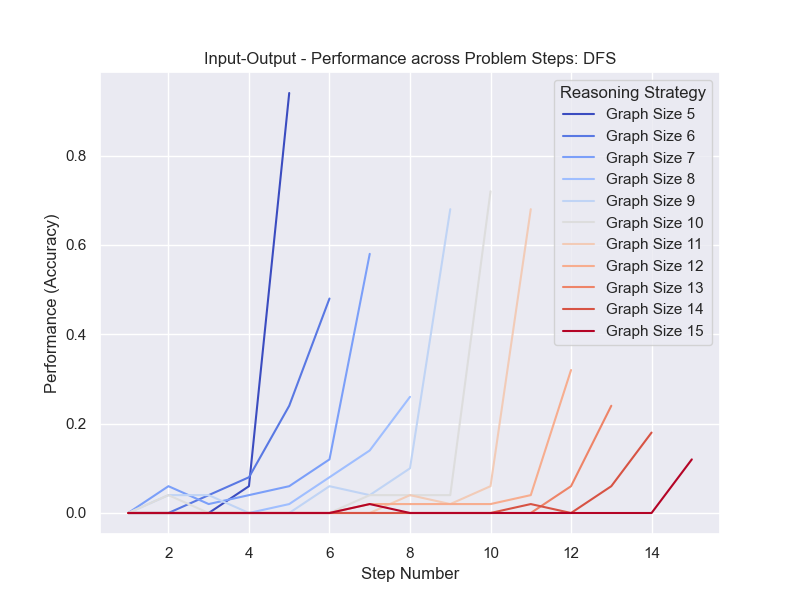}
        \caption{DFS - Input-Output}
    \end{subfigure}
    \hfill
    \begin{subfigure}{0.4\linewidth}
        \centering
        \includegraphics[width=\linewidth]{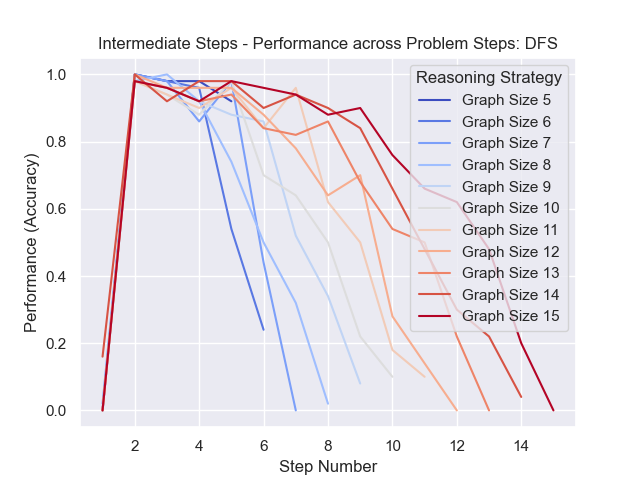}
        \caption{DFS - Intermediate Steps}
    \end{subfigure}

    % Row 3: Dijkstra
    \begin{subfigure}{0.4\linewidth}
        \centering
        \includegraphics[width=\linewidth]{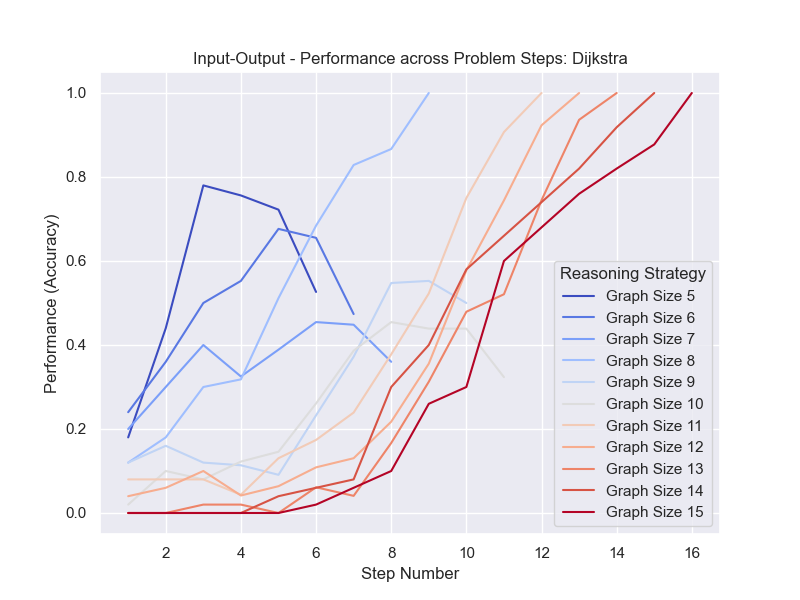}
        \caption{Dijkstra - Input-Output}
    \end{subfigure}
    \hfill
    \begin{subfigure}{0.4\linewidth}
        \centering
        \includegraphics[width=\linewidth]{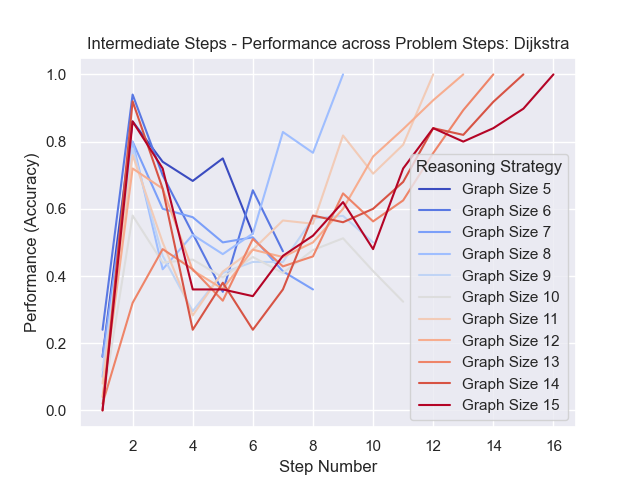}
        \caption{Dijkstra - Intermediate Steps}
    \end{subfigure}

    % Row 4: Floyd-Warshall
    \begin{subfigure}{0.4\linewidth}
        \centering
        \includegraphics[width=\linewidth]{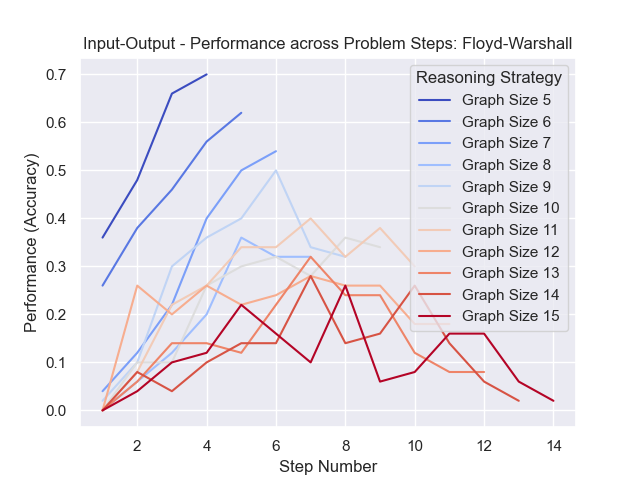}
        \caption{Floyd-Warshall - Input-Output}
    \end{subfigure}
    \hfill
    \begin{subfigure}{0.4\linewidth}
        \centering
        \includegraphics[width=\linewidth]{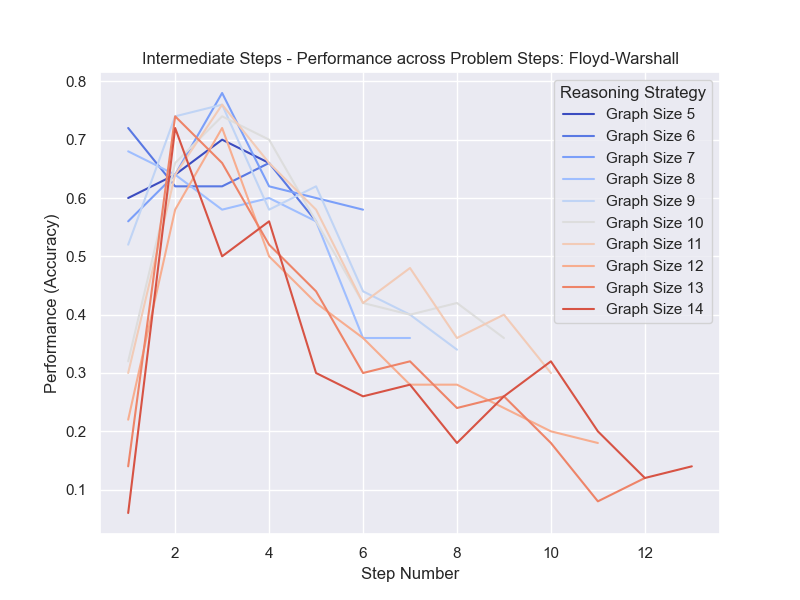}
        \caption{Floyd-Warshall - Intermediate Steps}
    \end{subfigure}

    % Row 5: Prim's MST
    \begin{subfigure}{0.4\linewidth}
        \centering
        \includegraphics[width=\linewidth]{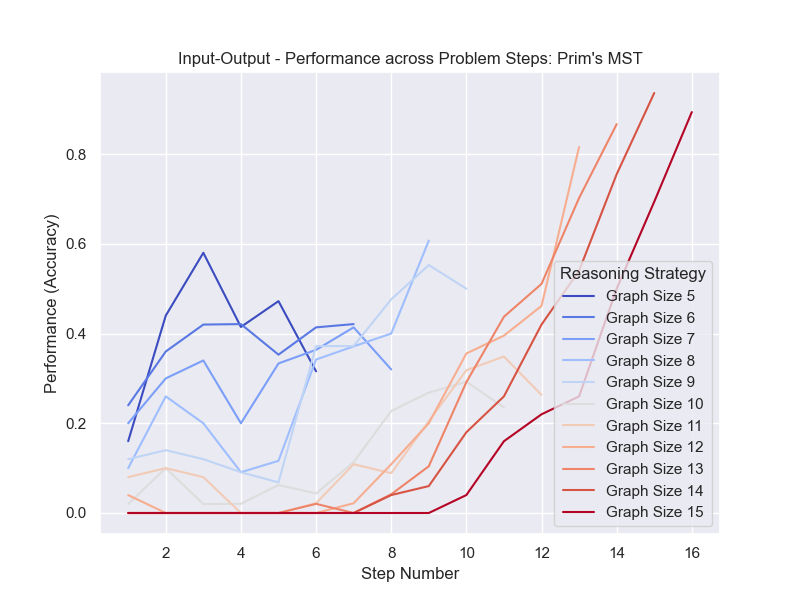}
        \caption{Prim's MST - Input-Output}
    \end{subfigure}
    \hfill
    \begin{subfigure}{0.4\linewidth}
        \centering
        \includegraphics[width=\linewidth]{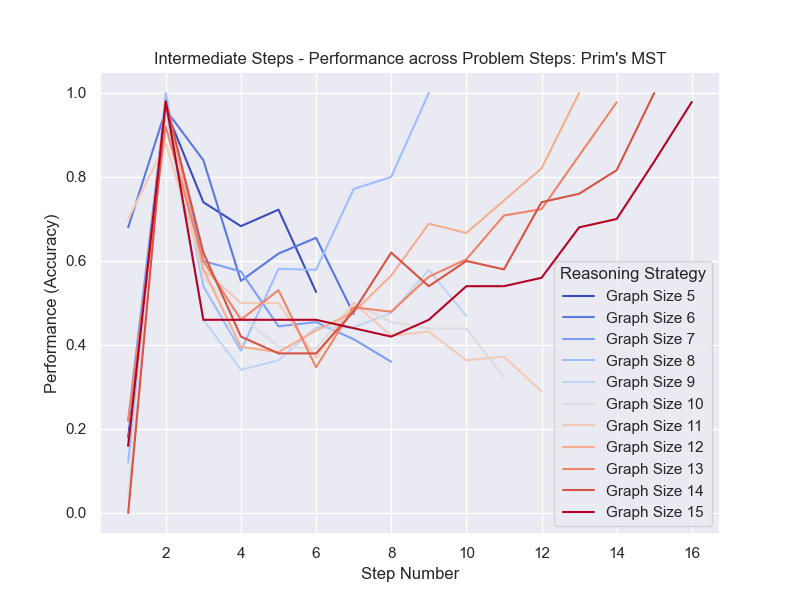}
        \caption{Prim's MST - Intermediate Steps}
    \end{subfigure}

\caption{Line plots showing the performance on intermediate steps of each model across algorithm and graph size.}
\label{fig:scatterplots_all_algorithms}
\end{figure*}

\newpage
\subsection{Full Results}
\begin{table*}[h!]
\centering
\resizebox{\textwidth}{!}{
\begin{tabular}{|c|c|c|c|c|c|c|c|c|c|c|c|c|c|c|c|}
\hline
Algorithm & Prompting Strategy & Model & 5 & 6 & 7 & 8 & 9 & 10 & 11 & 12 & 13 & 14 & 15 & 20 & 50 \\ \hline
Prim's MST & Input-Output (Output + Reasoning) & Mistral-Instruct & 26.07 & 27.37 & 41.44 & 27.61 & 30.86 & 19.64 & 18.55 & 7.63 & 11.73 & 7.19 & 5.78 & 5.79 & 0.00 \\ \hline
Prim's MST & Input-Output (Output + Reasoning) & Mistral & 8.27 & 14.50 & 51.04 & 29.77 & 38.06 & 12.63 & 13.36 & 13.82 & 12.33 & 13.99 & 5.35 & 2.45 & 2.48 \\ \hline
Prim's MST & Input-Output (Output + Reasoning) & Llama-Instruct & 20.80 & 17.30 & 14.06 & 8.90 & 9.66 & 7.07 & 2.65 & 4.62 & 3.59 & 2.94 & 2.10 & 1.98 & 0.00 \\ \hline
Prim's MST & Input-Output (Output + Reasoning) & Llama & 29.07 & 34.60 & 12.79 & 17.12 & 0.00 & 5.64 & 20.63 & 17.33 & 32.68 & 33.01 & 32.30 & 20.37 & 20.39 \\ \hline
Prim's MST & Input-Output (Output + Reasoning) & GPT-4o & 12.56 & 10.95 & 4.58 & 5.01 & 3.03 & 1.51 & 1.17 & 0.00 & 0.78 & 0.11 & 0.22 & 0.00 & 0.00 \\ \hline
Prim's MST & Intermediate Steps & Mistral-Instruct & 53.90 & 55.20 & 50.25 & 41.90 & 37.20 & 0.00 & 24.50 & 23.76 & 26.41 & 22.84 & 18.10 & 18.11 & 11.74 \\ \hline
Prim's MST & Intermediate Steps & Mistral & 72.00 & 41.20 & 53.35 & 39.26 & 33.85 & 0.00 & 30.63 & 25.22 & 14.60 & 22.31 & 15.85 & 11.30 & 10.88 \\ \hline
Prim's MST & Intermediate Steps & Llama-Instruct & 63.37 & 74.10 & 57.43 & 61.02 & 49.98 & 48.09 & 54.17 & 58.13 & 58.36 & 56.46 & 54.26 & 44.05 & 44.06 \\ \hline
Prim's MST & Intermediate Steps & Llama & 84.10 & 78.73 & 58.65 & 68.17 & 44.26 & 0.00 & 45.12 & 52.69 & 52.58 & 56.78 & 54.06 & 42.81 & 42.82 \\ \hline
Prim's MST & Input-Output (Output Only) & Llama-Instruct & 43.40 & 43.23 & 37.10 & 26.83 & 28.07 & 11.20 & 16.27 & 12.45 & 15.19 & 15.78 & 12.10 & 9.67 & 6.87 \\ \hline
Prim's MST & Input-Output (Output Only) & Mistral-Instruct & 62.67 & 47.23 & 52.46 & 39.75 & 38.89 & 33.50 & 29.67 & 22.63 & 22.82 & 24.89 & 20.51 & 11.38 & 11.38 \\ \hline
Prim's MST & Input-Output (Output Only) & Mistral & 75.33 & 38.20 & 58.52 & 39.37 & 33.28 & 31.79 & 30.52 & 26.03 & 12.23 & 20.75 & 19.81 & 17.26 & 9.78 \\ \hline
Prim's MST & Input-Output (Output Only) & Llama-Instruct & 34.02 & 44.21 & 36.22 & 36.15 & 34.38 & 14.49 & 14.96 & 10.12 & 10.29 & 40.67 & 39.75 & 38.59 & 27.23 \\ \hline
Prim's MST & Input-Output (Output Only) & Llama & 87.60 & 70.20 & 65.98 & 61.15 & 0.00 & 51.71 & 56.60 & 43.08 & 53.49 & 58.03 & 58.23 & 56.18 & 56.18 \\ \hline
Floyd-Warshall & Input-Output (Output + Reasoning) & Mistral-Instruct & 0.00 & 0.00 & 0.00 & 0.00 & 0.00 & 0.00 & 0.00 & 0.00 & 18.00 & 10.83 & 9.23 & 5.01 & 5.01 \\ \hline
Floyd-Warshall & Input-Output (Output + Reasoning) & Mistral & 0.00 & 0.00 & 0.00 & 0.00 & 0.00 & 0.00 & 0.00 & 0.00 & 18.18 & 16.33 & 12.92 & 11.71 & 10.52 \\ \hline
Floyd-Warshall & Input-Output (Output + Reasoning) & Llama-Instruct & 40.77 & 41.20 & 39.60 & 41.53 & 36.86 & 6.25 & 3.78 & 3.40 & 20.91 & 16.90 & 15.60 & 2.02 & 2.03 \\ \hline
Floyd-Warshall & Input-Output (Output + Reasoning) & Llama & 70.67 & 56.00 & 50.80 & 51.67 & 40.57 & 33.50 & 28.89 & 33.20 & 21.09 & 21.00 & 19.08 & 14.53 & 14.53 \\ \hline
Floyd-Warshall & Input-Output (Output + Reasoning) & GPT-4o & 19.47 & 12.60 & 7.68 & 1.87 & 0.80 & 0.60 & 0.00 & 0.00 & 0.00 & 0.00 & 0.12 & 0.00 & 0.00 \\ \hline
Floyd-Warshall & Intermediate Steps & Mistral-Instruct & 43.33 & 53.50 & 45.20 & 27.00 & 21.14 & 0.00 & 24.89 & 20.60 & 17.82 & 14.17 & 10.92 & 7.42 & 6.18 \\ \hline
Floyd-Warshall & Intermediate Steps & Mistral & 68.00 & 58.00 & 9.60 & 35.00 & 29.71 & 0.00 & 21.56 & 18.00 & 14.00 & 8.17 & 8.92 & 1.07 & 1.08 \\ \hline
Floyd-Warshall & Intermediate Steps & Llama-Instruct & 64.67 & 65.50 & 64.00 & 57.00 & 58.00 & 52.75 & 51.11 & 38.00 & 35.27 & 31.33 & 30.93 & 20.75 & 20.75 \\ \hline
Floyd-Warshall & Intermediate Steps & Llama & 53.33 & 70.50 & 58.40 & 65.67 & 59.43 & 0.00 & 52.67 & 42.80 & 34.55 & 30.33 & 24.46 & 24.49 & 23.75 \\ \hline
Floyd-Warshall & Input-Output (Output Only) & Llama-Instruct & 50.00 & 41.50 & 25.60 & 17.67 & 28.86 & 21.50 & 26.00 & 21.60 & 15.27 & 12.83 & 11.69 & 11.70 & 5.74 \\ \hline
Floyd-Warshall & Input-Output (Output Only) & Mistral-Instruct & 0.00 & 0.00 & 0.00 & 0.00 & 0.00 & 24.75 & 21.78 & 24.40 & 19.45 & 14.33 & 11.54 & 4.82 & 4.83 \\ \hline
Floyd-Warshall & Input-Output (Output Only) & Mistral & 0.00 & 0.00 & 0.00 & 0.00 & 0.00 & 23.50 & 17.33 & 20.20 & 14.18 & 8.67 & 9.38 & 3.09 & 0.00 \\ \hline
Floyd-Warshall & Input-Output (Output Only) & Llama-Instruct & 52.98 & 41.20 & 51.52 & 47.40 & 41.43 & 39.25 & 34.80 & 30.68 & 27.89 & 23.47 & 23.47 & 3.78 & 3.78 \\ \hline
Floyd-Warshall & Input-Output (Output Only) & Llama & 50.67 & 69.00 & 63.60 & 65.00 & 62.86 & 60.00 & 48.22 & 44.60 & 36.55 & 29.17 & 26.46 & 26.47 & 15.38 \\ \hline
Dijkstra & Input-Output (Output + Reasoning) & Mistral-Instruct & 44.00 & 28.50 & 27.63 & 28.22 & 15.15 & 21.07 & 15.81 & 21.52 & 17.27 & 20.58 & 21.76 & 7.32 & 7.32 \\ \hline
Dijkstra & Input-Output (Output + Reasoning) & Mistral & 2.80 & 7.77 & 32.89 & 7.01 & 21.46 & 28.43 & 19.00 & 33.61 & 27.13 & 24.70 & 15.37 & 14.59 & 11.11 \\ \hline
Dijkstra & Input-Output (Output + Reasoning) & Llama-Instruct & 30.27 & 32.82 & 26.59 & 20.43 & 9.44 & 3.92 & 4.37 & 3.79 & 1.01 & 1.82 & 1.31 & 0.60 & 0.00 \\ \hline
Dijkstra & Input-Output (Output + Reasoning) & Llama & 25.47 & 10.53 & 4.00 & 39.05 & 0.00 & 20.79 & 30.37 & 18.86 & 40.96 & 38.05 & 31.14 & 31.17 & 31.18 \\ \hline
Dijkstra & Input-Output (Output + Reasoning) & GPT-4o & 28.69 & 23.05 & 22.28 & 18.42 & 13.87 & 9.66 & 6.42 & 5.38 & 2.82 & 0.82 & 0.75 & 0.00 & 0.00 \\ \hline
Dijkstra & Intermediate Steps & Mistral-Instruct & 78.83 & 63.57 & 58.89 & 49.98 & 38.29 & 0.00 & 42.13 & 28.14 & 33.13 & 28.45 & 27.43 & 24.33 & 22.81 \\ \hline
Dijkstra & Intermediate Steps & Mistral & 68.47 & 54.37 & 50.91 & 38.31 & 34.93 & 0.00 & 45.50 & 14.37 & 26.98 & 26.70 & 25.62 & 19.00 & 11.94 \\ \hline
Dijkstra & Intermediate Steps & Llama-Instruct & 61.27 & 60.43 & 50.03 & 50.76 & 47.66 & 42.24 & 55.79 & 56.65 & 49.13 & 55.47 & 58.56 & 51.57 & 51.13 \\ \hline
Dijkstra & Intermediate Steps & Llama & 81.70 & 74.77 & 62.50 & 55.35 & 57.41 & 0.00 & 53.76 & 49.18 & 52.19 & 56.58 & 52.93 & 49.33 & 49.34 \\ \hline
Dijkstra & Input-Output (Output Only) & Llama-Instruct & 56.00 & 50.67 & 41.36 & 45.80 & 29.40 & 24.30 & 33.23 & 27.93 & 24.28 & 32.36 & 29.45 & 29.45 & 21.44 \\ \hline
Dijkstra & Input-Output (Output Only) & Mistral-Instruct & 78.47 & 52.00 & 64.74 & 53.16 & 35.35 & 37.36 & 35.25 & 30.72 & 33.44 & 27.17 & 28.57 & 28.55 & 16.20 \\ \hline
Dijkstra & Input-Output (Output Only) & Mistral & 72.67 & 51.67 & 52.24 & 41.60 & 31.96 & 37.14 & 32.65 & 12.01 & 30.99 & 28.48 & 25.51 & 25.52 & 20.21 \\ \hline
Dijkstra & Input-Output (Output Only) & Llama-Instruct & 32.77 & 35.55 & 24.37 & 23.00 & 19.22 & 11.01 & 12.35 & 11.21 & 9.89 & 7.77 & 7.16 & 5.69 & 5.67 \\ \hline
Dijkstra & Input-Output (Output Only) & Llama & 86.00 & 73.23 & 67.54 & 53.45 & 0.00 & 55.74 & 43.67 & 42.77 & 54.27 & 58.98 & 53.28 & 53.26 & 49.47 \\ \hline
DFS & Input-Output (Output + Reasoning) & Mistral-Instruct & 19.00 & 30.00 & 28.00 & 19.43 & 16.75 & 11.33 & 11.20 & 9.64 & 9.17 & 4.77 & 8.29 & 0.00 & 0.00 \\ \hline
DFS & Input-Output (Output + Reasoning) & Mistral & 42.00 & 32.00 & 13.67 & 26.29 & 20.50 & 8.44 & 8.40 & 5.64 & 9.00 & 7.54 & 12.29 & 7.07 & 0.00 \\ \hline
DFS & Input-Output (Output + Reasoning) & Llama-Instruct & 5.46 & 24.40 & 21.33 & 20.17 & 21.30 & 18.49 & 23.08 & 17.53 & 13.47 & 18.37 & 15.60 & 15.62 & 7.11 \\ \hline
DFS & Input-Output (Output + Reasoning) & Llama & 5.00 & 20.50 & 19.20 & 21.71 & 18.97 & 16.15 & 21.60 & 15.84 & 12.36 & 17.10 & 14.00 & 15.00 & 7.12 \\ \hline
DFS & Input-Output (Output + Reasoning) & GPT-4o & 0.50 & 1.60 & 0.80 & 0.23 & 0.34 & 0.36 & 0.18 & 0.22 & 0.07 & 0.07 & 0.06 & 0.00 & 0.00 \\ \hline
DFS & Intermediate Steps & Mistral-Instruct & 82.50 & 56.00 & 47.67 & 54.57 & 58.75 & 0.00 & 58.20 & 24.18 & 55.00 & 47.08 & 51.43 & 41.58 & 41.72 \\ \hline
DFS & Intermediate Steps & Mistral & 92.00 & 37.60 & 75.33 & 59.71 & 65.00 & 0.00 & 52.00 & 56.00 & 50.50 & 42.62 & 55.43 & 54.01 & 45.27 \\ \hline
DFS & Intermediate Steps & Llama-Instruct & 74.00 & 73.50 & 76.80 & 69.00 & 73.14 & 70.25 & 74.44 & 71.60 & 73.09 & 75.50 & 77.23 & 67.84 & 63.53 \\ \hline
DFS & Intermediate Steps & Llama & 25.00 & 21.20 & 44.00 & 37.43 & 26.50 & 0.00 & 59.60 & 52.18 & 51.67 & 62.31 & 58.43 & 49.95 & 49.99 \\ \hline
DFS & Input-Output (Output Only) & Llama-Instruct & 1.50 & 7.20 & 5.00 & 3.43 & 3.50 & 1.78 & 1.20 & 1.09 & 0.50 & 0.62 & 0.14 & 0.00 & 0.00 \\ \hline
DFS & Input-Output (Output Only) & Mistral-Instruct & 85.00 & 59.20 & 48.67 & 63.71 & 58.50 & 51.33 & 59.40 & 23.64 & 53.17 & 48.31 & 46.00 & 42.27 & 42.32 \\ \hline
DFS & Input-Output (Output Only) & Mistral & 93.00 & 41.20 & 68.33 & 60.29 & 61.50 & 55.33 & 54.60 & 58.18 & 49.33 & 42.62 & 56.71 & 51.42 & 51.46 \\ \hline
DFS & Input-Output (Output Only) & Llama-Instruct & 27.33 & 27.28 & 37.87 & 31.71 & 33.70 & 40.04 & 47.80 & 45.31 & 38.33 & 35.02 & 43.60 & 13.52 & 13.54 \\ \hline
DFS & Input-Output (Output Only) & Llama & 25.00 & 18.40 & 35.67 & 46.57 & 31.25 & 26.89 & 57.40 & 52.55 & 48.50 & 65.54 & 57.43 & 56.38 & 56.40 \\ \hline
BFS & Input-Output (Output + Reasoning) & Mistral-Instruct & 0.00 & 31.58 & 39.15 & 18.83 & 15.48 & 2.02 & 5.07 & 7.35 & 5.24 & 1.43 & 3.47 & 3.47 & 1.40 \\ \hline
BFS & Input-Output (Output + Reasoning) & Mistral & 0.67 & 27.59 & 32.31 & 19.79 & 5.95 & 6.45 & 1.06 & 0.00 & 0.00 & 2.21 & 3.19 & 0.00 & 0.00 \\ \hline
BFS & Input-Output (Output + Reasoning) & Llama-Instruct & 0.00 & 0.00 & 0.00 & 0.00 & 0.00 & 0.00 & 0.00 & 0.00 & 0.00 & 0.00 & 0.00 & 0.00 & 1.73 \\ \hline
BFS & Input-Output (Output + Reasoning) & Llama & 32.29 & 5.79 & 3.50 & 3.37 & 0.00 & 11.39 & 0.00 & 3.03 & 13.84 & 11.53 & 7.53 & 7.53 & 4.69 \\ \hline
BFS & Input-Output (Output + Reasoning) & GPT-4o & 12.50 & 12.27 & 11.70 & 12.69 & 13.11 & 10.89 & 12.78 & 11.78 & 7.96 & 7.18 & 8.92 & 0.00 & 0.00 \\ \hline
BFS & Intermediate Steps & Mistral-Instruct & 61.67 & 52.75 & 28.13 & 38.13 & 28.48 & 0.00 & 33.77 & 40.07 & 21.39 & 24.89 & 34.83 & 28.27 & 19.88 \\ \hline
BFS & Intermediate Steps & Mistral & 53.33 & 76.92 & 45.04 & 58.29 & 45.80 & 0.00 & 44.06 & 44.75 & 30.03 & 27.67 & 3.33 & 3.37 & 3.39 \\ \hline
BFS & Intermediate Steps & Llama-Instruct & 93.33 & 88.54 & 79.80 & 89.39 & 75.00 & 74.01 & 71.92 & 66.31 & 59.65 & 54.23 & 38.57 & 38.57 & 38.57 \\ \hline
BFS & Intermediate Steps & Llama & 42.67 & 0.00 & 65.28 & 56.18 & 60.23 & 0.00 & 81.05 & 54.65 & 59.83 & 52.57 & 56.87 & 55.20 & 54.62 \\ \hline
BFS & Input-Output (Output Only) & Llama-Instruct & 42.67 & 44.04 & 29.44 & 33.86 & 19.02 & 31.65 & 28.01 & 28.72 & 29.44 & 27.91 & 28.93 & 18.08 & 18.09 \\ \hline
BFS & Input-Output (Output Only) & Mistral-Instruct & 0.00 & 47.41 & 52.99 & 38.33 & 32.01 & 39.73 & 40.25 & 30.20 & 23.37 & 28.91 & 29.31 & 20.09 & 20.13 \\ \hline
BFS & Input-Output (Output Only) & Mistral & 72.92 & 73.07 & 56.32 & 62.38 & 46.84 & 32.61 & 47.09 & 45.61 & 37.07 & 30.58 & 2.55 & 2.56 & 2.56 \\ \hline
BFS & Input-Output (Output Only) & Llama-Instruct & 13.29 & 20.98 & 29.99 & 30.97 & 25.68 & 18.47 & 22.72 & 21.16 & 17.60 & 19.47 & 14.49 & 6.61 & 11.60 \\ \hline
BFS & Input-Output (Output Only) & Llama & 35.76 & 0.00 & 66.41 & 53.76 & 63.84 & 48.96 & 85.89 & 62.86 & 58.44 & 57.79 & 49.54 & 49.62 & 49.67 \\ \hline
\end{tabular}}
\caption{Full Intermediate Steps Exact Accuracy Results}
\end{table*}

\begin{table*}[h!]
\centering
\resizebox{\textwidth}{!}{
\begin{tabular}{|c|c|c|c|c|c|c|c|c|c|c|c|c|c|c|c|}
\hline
Algorithm & Prompting Strategy & Model & 5 & 6 & 7 & 8 & 9 & 10 & 11 & 12 & 13 & 14 & 15 & 20 & 50 \\ \hline
Prim's MST & Input-Output (Output + Reasoning) & Mistral-Instruct & 30.80 & 33.11 & 44.28 & 29.44 & 33.46 & 21.30 & 20.03 & 7.43 & 12.33 & 7.92 & 5.92 & 0.00 & 0.00 \\ \hline
Prim's MST & Input-Output (Output + Reasoning) & Mistral & 21.40 & 15.65 & 53.29 & 32.97 & 39.95 & 16.44 & 16.75 & 15.02 & 12.16 & 13.99 & 5.01 & 0.67 & 0.69 \\ \hline
Prim's MST & Input-Output (Output + Reasoning) & Llama-Instruct & 26.79 & 21.96 & 17.41 & 11.86 & 11.53 & 9.20 & 3.53 & 5.17 & 3.65 & 2.97 & 2.32 & 2.02 & 0.00 \\ \hline
Prim's MST & Input-Output (Output + Reasoning) & Llama & 26.07 & 41.22 & 10.95 & 18.95 & 0.00 & 6.80 & 22.83 & 18.72 & 37.49 & 37.10 & 35.32 & 29.03 & 28.79 \\ \hline
Prim's MST & Input-Output (Output + Reasoning) & GPT-4o & 13.20 & 11.85 & 4.83 & 5.25 & 3.24 & 1.56 & 1.14 & 0.00 & 0.83 & 0.11 & 0.20 & 0.00 & 0.00 \\ \hline
Prim's MST & Intermediate Steps & Mistral-Instruct & 56.57 & 55.87 & 52.19 & 45.59 & 38.81 & 0.00 & 25.80 & 25.70 & 30.64 & 25.47 & 20.37 & 18.07 & 18.08 \\ \hline
Prim's MST & Intermediate Steps & Mistral & 71.77 & 42.84 & 55.62 & 40.47 & 38.63 & 0.00 & 31.29 & 27.10 & 14.70 & 24.17 & 17.87 & 13.42 & 13.44 \\ \hline
Prim's MST & Intermediate Steps & Llama-Instruct & 67.23 & 75.85 & 59.64 & 65.90 & 52.00 & 49.03 & 53.61 & 61.63 & 61.21 & 59.23 & 56.87 & 56.87 & 53.92 \\ \hline
Prim's MST & Intermediate Steps & Llama & 83.87 & 79.20 & 60.85 & 71.84 & 44.60 & 0.00 & 46.33 & 53.35 & 55.63 & 59.68 & 56.95 & 51.66 & 45.11 \\ \hline
Prim's MST & Input-Output (Output Only) & Llama-Instruct & 46.03 & 46.11 & 40.31 & 31.19 & 31.20 & 13.96 & 18.13 & 16.71 & 19.97 & 20.88 & 17.00 & 17.00 & 13.28 \\ \hline
Prim's MST & Input-Output (Output Only) & Mistral-Instruct & 65.13 & 52.23 & 54.22 & 40.60 & 41.42 & 38.09 & 32.89 & 23.21 & 26.48 & 26.84 & 23.73 & 17.59 & 17.60 \\ \hline
Prim's MST & Input-Output (Output Only) & Mistral & 74.27 & 40.01 & 61.01 & 39.57 & 37.90 & 34.94 & 33.93 & 27.67 & 12.65 & 23.12 & 23.38 & 15.53 & 15.20 \\ \hline
Prim's MST & Input-Output (Output Only) & Llama-Instruct & 38.47 & 52.27 & 39.88 & 39.88 & 38.00 & 14.56 & 14.89 & 9.91 & 9.75 & 44.40 & 43.75 & 41.41 & 41.41 \\ \hline
Prim's MST & Input-Output (Output Only) & Llama & 86.40 & 72.50 & 67.71 & 61.56 & 0.00 & 56.42 & 60.31 & 42.56 & 56.81 & 61.11 & 60.86 & 53.58 & 53.62 \\ \hline
Floyd-Warshall & Input-Output (Output + Reasoning) & Mistral-Instruct & 0.00 & 0.00 & 0.00 & 0.00 & 0.00 & 0.00 & 0.00 & 0.00 & 17.50 & 10.15 & 9.14 & 9.14 & 0.00 \\ \hline
Floyd-Warshall & Input-Output (Output + Reasoning) & Mistral & 0.00 & 0.00 & 0.00 & 0.00 & 0.00 & 0.00 & 0.00 & 0.00 & 17.83 & 15.38 & 12.57 & 12.57 & 3.62 \\ \hline
Floyd-Warshall & Input-Output (Output + Reasoning) & Llama-Instruct & 46.43 & 42.64 & 40.53 & 40.34 & 36.00 & 5.87 & 3.44 & 3.09 & 20.60 & 15.94 & 14.60 & 1.92 & 1.92 \\ \hline
Floyd-Warshall & Input-Output (Output + Reasoning) & Llama & 71.00 & 55.60 & 48.00 & 48.86 & 41.00 & 31.11 & 28.00 & 32.00 & 20.83 & 19.38 & 17.86 & 6.47 & 6.47 \\ \hline
Floyd-Warshall & Input-Output (Output + Reasoning) & GPT-4o & 20.00 & 12.48 & 7.47 & 2.06 & 0.80 & 0.62 & 0.00 & 0.00 & 0.00 & 0.00 & 0.11 & 0.00 & 0.00 \\ \hline
Floyd-Warshall & Intermediate Steps & Mistral-Instruct & 48.00 & 52.40 & 45.67 & 26.86 & 20.75 & 0.00 & 23.40 & 19.09 & 16.67 & 13.38 & 10.29 & 4.34 & 0.00 \\ \hline
Floyd-Warshall & Intermediate Steps & Mistral & 66.50 & 56.40 & 9.33 & 33.71 & 29.00 & 0.00 & 20.20 & 17.09 & 13.00 & 7.69 & 8.57 & 6.26 & 6.27 \\ \hline
Floyd-Warshall & Intermediate Steps & Llama-Instruct & 65.00 & 63.60 & 63.00 & 54.00 & 55.00 & 50.89 & 49.00 & 36.18 & 33.33 & 30.00 & 28.51 & 28.51 & 16.37 \\ \hline
Floyd-Warshall & Intermediate Steps & Llama & 41.50 & 66.40 & 57.67 & 60.86 & 56.00 & 0.00 & 50.00 & 40.55 & 32.67 & 28.62 & 22.86 & 22.88 & 22.22 \\ \hline
Floyd-Warshall & Input-Output (Output Only) & Llama-Instruct & 55.00 & 45.60 & 30.33 & 19.71 & 29.25 & 22.89 & 26.40 & 21.27 & 14.67 & 12.00 & 11.00 & 0.99 & 1.00 \\ \hline
Floyd-Warshall & Input-Output (Output Only) & Mistral-Instruct & 0.00 & 0.00 & 0.00 & 0.00 & 0.00 & 23.56 & 20.80 & 22.91 & 18.50 & 13.54 & 11.00 & 10.42 & 1.64 \\ \hline
Floyd-Warshall & Input-Output (Output Only) & Mistral & 0.00 & 0.00 & 0.00 & 0.00 & 0.00 & 21.78 & 16.40 & 19.27 & 13.67 & 8.15 & 8.71 & 6.74 & 0.00 \\ \hline
Floyd-Warshall & Input-Output (Output Only) & Llama-Instruct & 57.14 & 43.04 & 50.93 & 45.37 & 40.05 & 38.31 & 33.76 & 29.60 & 26.83 & 22.34 & 22.34 & 3.58 & 3.58 \\ \hline
Floyd-Warshall & Input-Output (Output Only) & Llama & 41.00 & 65.60 & 61.67 & 61.43 & 60.75 & 55.33 & 46.20 & 42.36 & 35.00 & 27.38 & 24.71 & 24.72 & 24.24 \\ \hline
Dijkstra & Input-Output (Output + Reasoning) & Mistral-Instruct & 49.33 & 37.50 & 34.63 & 33.19 & 15.26 & 25.24 & 17.62 & 22.39 & 16.90 & 21.24 & 22.33 & 20.06 & 8.39 \\ \hline
Dijkstra & Input-Output (Output + Reasoning) & Mistral & 11.60 & 13.03 & 38.73 & 10.55 & 22.59 & 31.15 & 17.95 & 34.50 & 29.05 & 25.73 & 15.96 & 16.04 & 16.10 \\ \hline
Dijkstra & Input-Output (Output + Reasoning) & Llama-Instruct & 30.42 & 32.83 & 25.61 & 23.43 & 13.34 & 6.59 & 6.52 & 5.49 & 1.53 & 2.67 & 1.83 & 0.91 & 0.00 \\ \hline
Dijkstra & Input-Output (Output + Reasoning) & Llama & 20.33 & 14.43 & 2.67 & 44.16 & 0.00 & 22.13 & 30.86 & 19.04 & 44.04 & 40.60 & 33.98 & 34.01 & 34.02 \\ \hline
Dijkstra & Input-Output (Output + Reasoning) & GPT-4o & 30.25 & 24.37 & 24.00 & 19.55 & 14.74 & 10.47 & 6.48 & 5.53 & 2.79 & 0.77 & 0.71 & 0.40 & 0.00 \\ \hline
Dijkstra & Intermediate Steps & Mistral-Instruct & 78.43 & 65.25 & 60.04 & 52.13 & 41.43 & 0.00 & 45.67 & 30.30 & 35.95 & 30.15 & 29.48 & 22.63 & 22.65 \\ \hline
Dijkstra & Intermediate Steps & Mistral & 70.63 & 55.92 & 51.30 & 42.83 & 37.98 & 0.00 & 48.40 & 13.94 & 30.10 & 29.74 & 27.80 & 17.09 & 17.12 \\ \hline
Dijkstra & Intermediate Steps & Llama-Instruct & 64.97 & 62.29 & 54.63 & 57.66 & 50.10 & 43.17 & 59.20 & 60.11 & 52.99 & 58.46 & 61.17 & 56.35 & 56.35 \\ \hline
Dijkstra & Intermediate Steps & Llama & 82.10 & 73.84 & 61.37 & 60.30 & 58.70 & 0.00 & 53.35 & 50.86 & 53.51 & 59.34 & 55.89 & 46.38 & 46.39 \\ \hline
Dijkstra & Input-Output (Output Only) & Llama-Instruct & 60.30 & 55.73 & 45.56 & 52.40 & 32.97 & 27.67 & 38.79 & 33.28 & 29.66 & 36.91 & 33.89 & 33.88 & 27.20 \\ \hline
Dijkstra & Input-Output (Output Only) & Mistral-Instruct & 79.27 & 53.29 & 65.31 & 54.45 & 38.68 & 40.97 & 35.44 & 31.48 & 35.20 & 28.86 & 30.96 & 21.85 & 21.86 \\ \hline
Dijkstra & Input-Output (Output Only) & Mistral & 73.93 & 52.78 & 53.32 & 43.18 & 36.77 & 42.30 & 34.01 & 12.62 & 32.64 & 31.26 & 27.59 & 20.67 & 16.36 \\ \hline
Dijkstra & Input-Output (Output Only) & Llama-Instruct & 35.30 & 39.92 & 25.39 & 23.84 & 19.81 & 11.77 & 13.59 & 11.94 & 9.68 & 8.17 & 7.05 & 5.55 & 0.00 \\ \hline
Dijkstra & Input-Output (Output Only) & Llama & 85.40 & 73.75 & 61.90 & 53.88 & 0.00 & 59.95 & 41.73 & 41.83 & 54.68 & 61.74 & 56.06 & 51.71 & 51.71 \\ \hline
DFS & Input-Output (Output + Reasoning) & Mistral-Instruct & 18.40 & 25.67 & 24.57 & 18.50 & 15.33 & 10.20 & 10.36 & 9.00 & 8.46 & 4.86 & 7.87 & 0.00 & 0.00 \\ \hline
DFS & Input-Output (Output + Reasoning) & Mistral & 35.20 & 28.67 & 11.71 & 23.25 & 18.22 & 7.60 & 8.18 & 5.33 & 8.46 & 7.29 & 11.73 & 4.14 & 0.00 \\ \hline
DFS & Input-Output (Output + Reasoning) & Llama-Instruct & 9.33 & 23.73 & 18.86 & 18.05 & 19.64 & 17.52 & 21.71 & 16.43 & 12.98 & 17.80 & 15.09 & 15.04 & 7.16 \\ \hline
DFS & Input-Output (Output + Reasoning) & Llama & 4.00 & 23.73 & 18.86 & 20.25 & 19.64 & 17.52 & 21.71 & 16.43 & 12.98 & 17.80 & 15.09 & 15.04 & 7.16 \\ \hline
DFS & Input-Output (Output + Reasoning) & GPT-4o & 11.20 & 6.80 & 3.89 & 3.20 & 2.22 & 1.84 & 2.11 & 1.47 & 1.11 & 1.26 & 0.80 & 0.00 & 0.40 \\ \hline
DFS & Intermediate Steps & Mistral-Instruct & 83.60 & 47.33 & 42.57 & 50.25 & 53.56 & 0.00 & 53.45 & 22.33 & 51.23 & 43.71 & 48.13 & 39.90 & 39.33 \\ \hline
DFS & Intermediate Steps & Mistral & 90.00 & 33.00 & 66.29 & 53.00 & 63.56 & 0.00 & 49.27 & 51.83 & 47.38 & 40.00 & 51.73 & 49.67 & 41.62 \\ \hline
DFS & Intermediate Steps & Llama-Instruct & 77.60 & 62.00 & 61.14 & 56.00 & 61.56 & 59.40 & 63.45 & 60.83 & 63.54 & 66.57 & 68.27 & 61.62 & 61.62 \\ \hline
DFS & Intermediate Steps & Llama & 20.00 & 18.33 & 37.71 & 32.75 & 23.56 & 0.00 & 54.18 & 47.83 & 47.85 & 58.14 & 54.53 & 48.71 & 43.90 \\ \hline
DFS & Input-Output (Output Only) & Llama-Instruct & 20.00 & 14.00 & 12.57 & 6.25 & 10.67 & 8.80 & 7.27 & 3.67 & 2.31 & 1.86 & 0.93 & 0.00 & 0.00 \\ \hline
DFS & Input-Output (Output Only) & Mistral-Instruct & 77.60 & 50.67 & 43.43 & 57.50 & 53.56 & 47.20 & 55.64 & 21.83 & 49.38 & 45.00 & 43.07 & 35.78 & 35.85 \\ \hline
DFS & Input-Output (Output Only) & Mistral & 81.60 & 36.00 & 59.14 & 54.25 & 59.78 & 54.20 & 51.82 & 53.67 & 45.85 & 40.29 & 53.20 & 40.22 & 40.26 \\ \hline
DFS & Input-Output (Output Only) & Llama-Instruct & 23.01 & 23.00 & 33.20 & 28.30 & 30.67 & 36.96 & 44.11 & 42.00 & 36.15 & 32.69 & 40.88 & 12.88 & 12.90 \\ \hline
DFS & Input-Output (Output Only) & Llama & 20.00 & 16.00 & 30.57 & 41.25 & 27.78 & 24.20 & 52.36 & 48.17 & 45.08 & 60.86 & 53.60 & 47.07 & 47.09 \\ \hline
BFS & Input-Output (Output + Reasoning) & Mistral-Instruct & 16.95 & 58.23 & 58.83 & 28.30 & 38.07 & 28.25 & 25.37 & 17.12 & 26.97 & 23.53 & 22.52 & 17.96 & 17.98 \\ \hline
BFS & Input-Output (Output + Reasoning) & Mistral & 16.76 & 50.63 & 56.27 & 29.54 & 21.60 & 23.93 & 5.97 & 16.49 & 2.57 & 10.00 & 17.88 & 4.93 & 4.00 \\ \hline
BFS & Input-Output (Output + Reasoning) & Llama-Instruct & 0.00 & 0.00 & 0.00 & 0.00 & 0.00 & 0.00 & 0.21 & 0.00 & 0.20 & 0.00 & 0.00 & 0.10 & 19.83 \\ \hline
BFS & Input-Output (Output + Reasoning) & Llama & 24.80 & 12.40 & 13.77 & 1.90 & 0.00 & 16.36 & 0.00 & 7.82 & 30.20 & 33.17 & 9.50 & 9.50 & 9.50 \\ \hline
BFS & Input-Output (Output + Reasoning) & GPT-4o & 30.30 & 23.69 & 19.80 & 20.04 & 20.38 & 16.43 & 18.17 & 17.57 & 11.79 & 12.72 & 14.24 & 0.00 & 0.00 \\ \hline
BFS & Intermediate Steps & Mistral-Instruct & 63.83 & 65.77 & 30.23 & 41.80 & 27.74 & 0.00 & 47.93 & 47.30 & 32.93 & 38.98 & 48.10 & 45.71 & 43.92 \\ \hline
BFS & Intermediate Steps & Mistral & 44.33 & 84.90 & 38.33 & 61.13 & 48.68 & 0.00 & 52.30 & 57.38 & 42.83 & 32.98 & 3.00 & 3.11 & 3.20 \\ \hline
BFS & Intermediate Steps & Llama-Instruct & 93.50 & 93.93 & 81.50 & 90.47 & 73.34 & 77.79 & 80.17 & 74.93 & 69.53 & 62.92 & 40.26 & 40.27 & 40.28 \\ \hline
BFS & Intermediate Steps & Llama & 26.50 & 0.00 & 54.67 & 52.47 & 53.16 & 0.00 & 82.90 & 45.64 & 46.83 & 50.61 & 59.36 & 44.48 & 44.58 \\ \hline
BFS & Input-Output (Output Only) & Llama-Instruct & 62.00 & 67.40 & 49.30 & 58.10 & 33.99 & 45.97 & 47.63 & 47.60 & 47.00 & 43.34 & 44.78 & 39.71 & 35.37 \\ \hline
BFS & Input-Output (Output Only) & Mistral-Instruct & 0.00 & 47.63 & 45.67 & 31.67 & 31.70 & 45.22 & 47.29 & 44.28 & 36.97 & 41.70 & 41.38 & 29.39 & 29.47 \\ \hline
BFS & Input-Output (Output Only) & Mistral & 56.20 & 81.17 & 48.97 & 54.40 & 45.07 & 39.10 & 55.09 & 55.81 & 49.90 & 38.83 & 1.82 & 1.85 & 1.88 \\ \hline
BFS & Input-Output (Output Only) & Llama-Instruct & 16.62 & 35.67 & 39.97 & 39.16 & 33.17 & 24.77 & 30.31 & 31.32 & 25.61 & 29.02 & 22.67 & 11.74 & 31.43 \\ \hline
BFS & Input-Output (Output Only) & Llama & 26.13 & 0.00 & 63.13 & 42.92 & 54.83 & 39.82 & 83.60 & 49.81 & 45.60 & 58.53 & 51.08 & 46.83 & 47.01 \\ \hline
\end{tabular}}
\caption{Full Algorithmic Trajectory Exact Accuracy Results}
\end{table*}

\end{document}